# Stochastic Variance-reduced Gradient Descent for Low-rank Matrix Recovery from Linear Measurements

Xiao Zhang[*][‡] and Lingxiao Wang[†][‡] and Quanquan Gu[§]

## Abstract

We study the problem of estimating low-rank matrices from linear measurements (a.k.a., matrix sensing) through nonconvex optimization. We propose an efficient stochastic variance reduced gradient descent algorithm to solve a nonconvex optimization problem of matrix sensing. Our algorithm is applicable to both noisy and noiseless settings. In the case with noisy observations, we prove that our algorithm converges to the unknown low-rank matrix at a linear rate up to the minimax optimal statistical error. And in the noiseless setting, our algorithm is guaranteed to linearly converge to the unknown low-rank matrix and achieves exact recovery with optimal sample complexity. Most notably, the overall computational complexity of our proposed algorithm, which is defined as the iteration complexity times per iteration time complexity, is lower than the state-of-the-art algorithms based on gradient descent. Experiments on synthetic data corroborate the superiority of the proposed algorithm over the state-of-the-art algorithms.

## 1 Introduction

We consider the problem of matrix sensing (Recht et al., 2010; Negahban and Wainwright, 2011), where the aim is to recover the unknown rank-$r$ matrix $\mathbf{X}^* \in \mathbb{R}^{d_1 \times d_2}$ from linear measurements $\boldsymbol{y} = \mathcal{A}_N(\mathbf{X}^*) + \boldsymbol{\epsilon}$, where $\mathcal{A}_N : \mathbb{R}^{d_1 \times d_2} \to \mathbb{R}^N$ is a linear measurement operator such that $\mathcal{A}_N(\mathbf{X}^*) = (\langle \mathbf{A}_1, \mathbf{X}^* \rangle, \langle \mathbf{A}_2, \mathbf{X}^* \rangle, \ldots, \langle \mathbf{A}_N, \mathbf{X}^* \rangle)^\top$, where $\langle \mathbf{A}_i, \mathbf{X}^* \rangle$ denotes the trace inner product on matrix space, i.e., $\langle \mathbf{A}_i, \mathbf{X}^* \rangle := \text{Tr}(\mathbf{A}_i^\top \mathbf{X}^*)$, and each entry of the noise vector $\boldsymbol{\epsilon}$ follows i.i.d. sub-Gaussian distribution with parameter $\nu$. In particular, we call the random matrix $\mathbf{A}_i \in \mathbb{R}^{d_1 \times d_2}$ as the sensing matrix. Therefore, we are interested in solving the following nonconvex rank minimization problem with equality constraint

$$\min_{\mathbf{X} \in \mathbb{R}^{d_1 \times d_2}} \text{rank}(\mathbf{X}) \quad \text{subject to} \quad \mathcal{A}_N(\mathbf{X}) = \boldsymbol{y}. \tag{1.1}$$

In this paper, we particularly consider the case that each sensing matrix $\mathbf{A}_i \in \mathbb{R}^{d_1 \times d_2}$ is a random matrix from Gaussian ensemble such that $(\mathbf{A}_i)_{jj} \sim N(0, 2)$ and $(\mathbf{A}_i)_{jk} \sim N(0, 1)$ for $j \neq k$. Lots of studies have been proposed in order to solve problem (1.1) efficiently, among which the most popular method is the following nuclear norm relaxation based approach (Recht et al., 2010; Recht, 2011; Candès and Tao, 2010; Rohde et al., 2011; Koltchinskii et al., 2011; Negahban and Wainwright, 2011, 2012; Gui and Gu, 2015)

$$\min_{\mathbf{X} \in \mathbb{R}^{d_1 \times d_2}} \|\mathbf{X}\|_* \quad \text{subject to} \quad \mathcal{A}_N(\mathbf{X}) = \mathbf{y}. \tag{1.2}$$

---

[*]Department of Statistics, University of Virginia, Charlottesville, VA 22904, USA; e-mail:`xz7bc@virginia.edu`

[†]Department of Systems and Information Engineering, University of Virginia, Charlottesville, VA 22904, USA; e-mail: `lw4wr@virginia.edu`

[‡]Equal Contribution

[§]Department of Systems and Information Engineering, Department of Computer Science, University of Virginia, Charlottesville, VA 22904, USA; e-mail: `qg5w@virginia.edu`

Since nuclear norm is the tightest convex relaxation of matrix rank, (1.2) relaxes the nonconvex optimization problem (1.1) to a convex one. Recht et al. (2010) first proved theoretical guarantees of problem (1.2) under the restricted isometry property of the linear measurement operator $\mathcal{A}_N$. Then lots of algorithms (Recht et al., 2010; Jain et al., 2010; Lee and Bresler, 2010) were proposed. Although the convex relaxation based method can recover the unknown low-rank matrix $\mathbf{X}^*$ with good theoretical guarantees, the drawback is its computational limitations. For instance, in order to recover the unknown low-rank matrix, most of these approaches need to perform singular value decomposition at each iteration, which results in huge computational complexity, especially for large matrices. To address this computational weakness, recent work (Jain et al., 2013; Zheng and Lafferty, 2015; Tu et al., 2015) proposed to recover the unknown low-rank matrix $\mathbf{X}^*$ through nonconvex optimization. More specifically, they proposed to factorize the rank-$r$ matrix $\mathbf{X} \in \mathbb{R}^{d_1 \times d_2}$ as $\mathbf{X} = \mathbf{U}\mathbf{V}^\top$, where $\mathbf{U} \in \mathbb{R}^{d_1 \times r}$, $\mathbf{V} \in \mathbb{R}^{d_2 \times r}$. This matrix factorization guarantees the low-rankness of $\mathbf{X}$, thus one can instead solve the following nonconvex optimization problem

$$\min_{\substack{\mathbf{U} \in \mathbb{R}^{d_1 \times r} \\ \mathbf{V} \in \mathbb{R}^{d_2 \times r}}} \mathcal{L}(\mathbf{U}\mathbf{V}^\top) := \frac{1}{2N} \sum_{i=1}^{N} \left( \langle \mathbf{A}_i, \mathbf{U}\mathbf{V}^\top \rangle - y_i \right)^2, \tag{1.3}$$

where $N$ is the number of sensing matrices. Although matrix factorization makes the optimization problem nonconvex, the computational complexity is significantly decreased compared with the convex optimization problem (1.2). A line of research (Jain et al., 2013; Zhao et al., 2015; Chen and Wainwright, 2015; Zheng and Lafferty, 2015; Tu et al., 2015; Bhojanapalli et al., 2015; Park et al., 2016a,b; Wang et al., 2016) has been established to study different nonconvex optimization algorithms for solving (1.3).

However, the nonconvex methods mentioned above are based on gradient descent (Zhao et al., 2015; Chen and Wainwright, 2015; Zheng and Lafferty, 2015; Tu et al., 2015; Bhojanapalli et al., 2015; Park et al., 2016a,b; Wang et al., 2016) or alternating minimization (Jain et al., 2013; Zhao et al., 2015), which are computationally expensive, especially for large scale problems, because they need to evaluate the full gradient at each iteration. In order to address this computational limitation, we reformulate the objective in (1.3) as a sum of $n$ component functions as follows

$$\min_{\substack{\mathbf{U} \in \mathbb{R}^{d_1 \times r} \\ \mathbf{V} \in \mathbb{R}^{d_2 \times r}}} \mathcal{L}(\mathbf{U}\mathbf{V}^\top) := \frac{1}{n} \sum_{i=1}^{n} \ell_i(\mathbf{U}\mathbf{V}^\top), \tag{1.4}$$

where each component function is defined as

$$\ell_i(\mathbf{U}\mathbf{V}^\top) = \frac{1}{2b}\sum_{j=1}^{b}\left(\langle \mathbf{A}_{i_j}, \mathbf{U}\mathbf{V}^\top\rangle - y_{i_j}\right)^2. \tag{1.5}$$

Note that each component function $\ell_i(\mathbf{U}\mathbf{V}^\top)$ is associated with $b$ observations satisfying $N = nb$. More specifically, for each $\ell_i(\mathbf{U}\mathbf{V}^\top)$, we define the linear measurement operator $\mathcal{A}_b^i : \mathbb{R}^{d_1 \times d_2} \to \mathbb{R}^b$ as $\mathcal{A}_b^i(\mathbf{X}) = (\langle \mathbf{A}_{i_1}, \mathbf{X}\rangle, \langle \mathbf{A}_{i_2}, \mathbf{X}\rangle, \ldots, \langle \mathbf{A}_{i_b}, \mathbf{X}\rangle)^\top$, and the corresponding observations are $\mathbf{y}^i = (y_{i_1}, y_{i_2}, \ldots, y_{i_b})^\top$. It is easy to show that the new formulation (1.4) is equivalent to (1.3). Based on the new formulation (1.4), we propose the first provable accelerated stochastic gradient descent algorithm for matrix sensing, which adopts the idea of stochastic variance reduced gradient descent (Johnson and Zhang, 2013). Our algorithm enjoys lower iteration and computational complexity, while ensuring the optimal sample complexity compared with existing alternatives (See Table 1 for a detailed comparison). More specifically, our algorithm is applicable to the case with noisy observations and that with noiseless observations. For noisy observations, we prove that our algorithm converges to the unknown low-rank matrix at a linear rate up to the statistical error, which matches the minimax lower bound $O(rd'/N)$ (Negahban and Wainwright, 2011). While in the noiseless case, our algorithm achieves the optimal sample complexity $O(rd')$ (Recht et al., 2010; Tu et al., 2015; Wang et al., 2016). Most importantly, to achieve $\epsilon$ accuracy, the computational complexity of our algorithm is $O((Nd'^2 + \kappa^2 bd'^2)\log(1/\epsilon))$. Here $N$ is the number of observations, $\kappa = \sigma_1/\sigma_r$ is the condition number, where $\sigma_1$ and $\sigma_r$ correspond to the 1-st and $r$-th singular values of the unknown matrix $\mathbf{X}^*$, respectively, and $b$ is the number of observations for each component function in (1.5). If the condition number $\kappa < n$, where $n$ is the number of component functions, the computational complexity of our algorithm is lower than those of the state-of-the-art algorithms proposed by Tu et al. (2015) for noiseless case, and Wang et al. (2016) for noisy case. Thorough experiments demonstrate that the performance of our method is better than the state-of-the art gradient descent based approaches.

The remainder of this paper is organized as follows. We briefly review some related work and compare the proposed approach with existing methods in Section 2. In Section 3, we illustrate the proposed algorithms in detail. We present the theoretical guarantees of the proposed methods in Section 4, and provide the corresponding proofs in Section 5. Section 6 provides numerical results of some synthetic data sets. Finally, we present the conclusion in Section 7.

**Notation** The capital symbols such as $\mathbf{A}$ is used to denote matrices and $[d]$ is used to denote $\{1, 2, \ldots, d\}$. We use $\langle \mathbf{A}, \mathbf{B}\rangle = \text{Tr}(\mathbf{A}^\top \mathbf{B})$ to denote the inner product between two matrices. For any matrix $\mathbf{A} \in \mathbb{R}^{d_1 \times d_2}$, the $(i,j)$-th entry of $\mathbf{A}$ is denoted by $A_{ij}$. Denote $d' = \max\{d_1, d_2\}$ and the $\ell$-th largest singular value of $\mathbf{A}$ by $\sigma_\ell(\mathbf{A})$. For any matrix $\mathbf{A} \in \mathbb{R}^{(d_1+d_2) \times r}$, we let $\mathbf{A}_U$ and $\mathbf{A}_V$ to denote the top $d_1 \times r$ and bottom $d_2 \times r$ matrices of $\mathbf{A}$, respectively. Consider a $d$-dimensional vector $\mathbf{x} = [x_1, x_2, \cdots, x_d]^\top \in \mathbb{R}^d$, the $\ell_q$ vector norm of $\mathbf{x}$ is denoted by $\|\mathbf{x}\|_q = (\Sigma_{i=1}^d |x_i|^q)^{1/q}$ for $0 < q < \infty$. We use $\|\mathbf{A}\|_F, \|\mathbf{A}\|_2$ to denote the Frobenius norm and the spectral norm of matrix $\mathbf{A}$, respectively. Let $\|\mathbf{A}\|_* = \sum_{i=1}^r \sigma_i(\mathbf{A})$ be the nuclear norm of $\mathbf{A}$, where $r$ is the rank of $\mathbf{A}$. In addition, given two sequences $\{a_n\}$ and $\{b_n\}$, if there exists a constant $0 < C_1 < \infty$ such that $a_n \le C_1 b_n$, then we write $a_n = O(b_n)$. Finally, we write $a_n = \Omega(b_n)$ if there exists a constant $0 < C_2 < \infty$ such that $a_n \ge C_2 b_n$. Finally, we write $a_n \asymp b_n$ if there exist positive constants $c$ and $C$ such that $c \leq a_n/b_n \leq C$.

## 2 Related Work

In this section, we discuss existing studies that are relevant to our work.

As mentioned in the introduction, given the restricted isometry property of the linear measurement operator $\mathcal{A}_N$, Recht et al. (2010) proposed to recover the low-rank matrix through nuclear norm minimization (1.1). Later on, lots of algorithms for solving nonconvex optimization problem (1.3) have been proposed. For instance, Jain et al. (2013) studied the performance of alternating minimization for matrix sensing. They showed that, provided a desired initial solution, their method enjoys linear convergence rate under the restricted isometry property that is similar to Recht et al. (2010); Zhao et al. (2015); Chen and Wainwright (2015); Tu et al. (2015) and ours. However, the restricted isometry property they assumed is more stringent compared with Recht et al. (2010); Chen and Wainwright (2015); Tu et al. (2015) and our work. Besides, since their algorithm requires to solve the least squares problems, which are often ill-conditioned at each iteration, the performance of their method is limited in practice. Later, a more unified analysis were established by Zhao et al. (2015). They proved that a broad class of algorithms, which includes gradient-based and alternating minimization approaches, can recover the unknown low-rank matrix. However, similar to Jain et al. (2013), their method also based on the restricted isometry property which has a more stringent form compared with others. Recently, Zheng and Lafferty (2015) provided an analysis of gradient descent approach for matrix sensing. They showed that, under a appropriate initial solution, their method is guaranteed to converge to the global optimum at a linear rate. More recently, Tu et al. (2015) established an improved analysis of gradient descent approach for matrix sensing problem compared with Zheng and Lafferty (2015) by a more sophisticated initialization procedure. In the mean time, Chen and Wainwright (2015); Bhojanapalli et al. (2015); Park et al. (2016b); Wang et al. (2016) studied the general low-rank matrix estimation problem using (projected) gradient descent. However, only Chen and Wainwright (2015) and Wang et al. (2016) discussed the matrix sensing problem with noisy observations. In particular, Chen and Wainwright (2015) provided a projected gradient descent algorithm to recover the unknown low-rank matrix from the linear measurements with restricted isometry property. Wang et al. (2016) proposed a unified framework for nonconvex low-rank matrix estimation under the restricted strongly convex and smooth conditions, which covers both noisy and noiseless matrix sensing as special cases. We also notice that in order to get rid of the initialization procedure, Bhojanapalli et al. (2016); Park et al. (2016c) showed that, under the restricted isometry property, all local minima of the matrix factorization based nonconvex optimization are global minimum for matrix sensing.

However, as discussed before, the aforementioned methods for solving problem (1.3) are based on gradient descent, which is computationally expensive since they have to calculate the full gradient at each iteration. Therefore, for large data set, stochastic gradient descent is often used to decrease the computational complexity . At each iteration, we only need to sample one or a mini-batch of the $n$ component functions $l_i$ (Nemirovski et al., 2009; Lan, 2012). However, due to the variance in estimating the gradient by random sampling, stochastic gradient descent often has a sublinear convergence rate even when $\mathcal{L}$ is strongly convex and smooth. Therefore, various types of stochastic gradient descent algorithms with variance reduction technique (Schmidt et al., 2013; Johnson and Zhang, 2013; Konečný and Richtárik, 2013; Defazio et al., 2014b; Mairal, 2014; Defazio et al., 2014a; Shamir, 2015a,b; Garber and Hazan, 2015; Li et al., 2016; Garber et al., 2016; Chen and Gu, 2016) were proposed to accelerate stochastic gradient descent. Inspired by the idea of stochastic variance reduced gradient descent (Johnson and Zhang, 2013; Xiao and Zhang, 2014; Konečný et al., 2014), we proposed an accelerated stochastic gradient descent algorithm for nonconvex optimization problem (1.4) to get over the computational barrier incurred by gradient descent and ensures the linear rate of convergence. Most remarkably, for the nonconvex optimization problem of principal component analysis, Shamir (2015a,b) proposed and analyzed stochastic variance reduced power method, and Garber and Hazan (2015); Garber et al. (2016) proposed to first reduce the problem of principal component analysis to a sequence of linear systems by the technique of shift-and-inverse preconditioning, then solve the linear systems by applying the stochastic variance reduced gradient. For the nonconvex optimization of sparsity constrained statistical learning, Li et al. (2016); Chen and Gu (2016) proposed variance reduced stochastic gradient and randomized block coordinate descent algorithms respectively. For general nonconvex finite-sum optimization problem, Reddi et al. (2016); Allen-Zhu and Hazan (2016) proposed stochastic variance reduced gradient descent algorithms, which are guaranteed to converge to the stationary point at a sublinear rate. However, none of the above nonconvex optimization algorithms and analyses can be adapted to solve problem (1.4).

Table 1 summarizes the detailed comparison among our proposed algorithm and existing state-of-the-art algorithms for the problem of matrix sensing. The nuclear norm relaxation algorithm (Recht et al., 2010) has optimal sample complexity $O(rd' \log d')$ up to a logarithmic term, but is computationally expensive. For AltMin (Jain et al., 2013), it can converge to the optimum in only $O(\log(1/\epsilon))$ iterations but requires $O(Nr^2d'^2 + r^3d'^3)$ number of gradient evaluations per-iteration which is very expensive. In addition, the sample complexity for AltMin is $O(\kappa^4 r^3 d' \log d')$ which is also very large compared to other algorithms. For Alternating GD (Zhao et al., 2015) and Projected GD (Chen and Wainwright, 2015), they both require large sample complexity $O(r^3 d' \log d')$, and their convergence rates depend on high-degree polynomial of condition number $\kappa$ which make their methods inefficient. For the gradient descent algorithm proposed by (Zheng and Lafferty, 2015), it has lower computational complexity $O(N\kappa^2 r^2 d'^3 \log(1/\epsilon))$ compared to the previous algorithms, but requires larger sample complexity $O(\kappa^2 r^3 d' \log d')$ due to its initialization procedure. The state-of-the-art gradient descent methods for noiseless (Tu et al., 2015) and noisy case (Wang et al., 2016) have the optimal sample complexity $O(rd')$. However, the computational complexity of their methods is $O(N\kappa d'^2 \log(1/\epsilon))$, which can be large due to the multiplication of the sample size $N$ and the condition number $\kappa$.

## 3 The Proposed Algorithm

In this section, we present our stochastic variance reduced gradient descent algorithm for solving (1.4). It is obvious that the optimization problem (1.4) has multiple solutions. Therefore, in order to guarantee the optimal solution is unique, following Tu et al. (2015); Zheng and Lafferty (2016); Park et al. (2016b); Wang et al. (2016), we impose a regularization term $\|\mathbf{U}^\top\mathbf{U} - \mathbf{V}^\top\mathbf{V}\|_F^2$ to ensure

| Algorithm | Sample Complexity | Iteration Complexity | Computational Complexity |
|---|---|---|---|
| Nuclear norm (Recht et al., 2010) | $O(rd'\log d')$ | $O\left(\frac{1}{\sqrt{\epsilon}}\right)$ | $O\left(\frac{d'^3}{\sqrt{\epsilon}}\right)$ |
| AltMin (Jain et al., 2013) | $O(\kappa^4 r^3 d' \log d')$ | $O\left(\log\left(\frac{1}{\epsilon}\right)\right)$ | $O\left((Nr^2 d'^2 + r^3 d'^3)\log(\frac{1}{\epsilon})\right)$ |
| Alternating GD (Zhao et al., 2015) | $O(r^3 d' \log d')$ | $O\left(\kappa^4 \log\left(\frac{1}{\epsilon}\right)\right)$ | $O\left(N\kappa^4 d'^2 \log\left(\frac{1}{\epsilon}\right)\right)$ |
| Projected GD (Chen and Wainwright, 2015) | $O(r^3 d' \log d')$ | $O\left(\kappa^{10} \log\left(\frac{1}{\epsilon}\right)\right)$ | $O\left(N\kappa^{10} d'^2 \log\left(\frac{1}{\epsilon}\right)\right)$ |
| GD (Zheng and Lafferty, 2015) | $O(\kappa^2 r^3 d' \log d')$ | $O\left(\kappa^2 r^2 d' \log\left(\frac{1}{\epsilon}\right)\right)$ | $O\left(N\kappa^2 r^2 d'^3 \log\left(\frac{1}{\epsilon}\right)\right)$ |
| GD (Tu et al., 2015) | $O(rd')$ | $O\left(\kappa \log\left(\frac{1}{\epsilon}\right)\right)$ | $O\left(N\kappa d'^2 \log\left(\frac{1}{\epsilon}\right)\right)$ |
| GD (Wang et al., 2016) | $O(rd')$ | $O\left(\kappa \log\left(\frac{1}{\epsilon}\right)\right)$ | $O\left(N\kappa d'^2 \log\left(\frac{1}{\epsilon}\right)\right)$ |
| This paper | $O(rd')$ | $O\left(\log\left(\frac{1}{\epsilon}\right)\right)$ | $O\left((Nd'^2 + \kappa^2 b d'^2)\log\left(\frac{1}{\epsilon}\right)\right)$ |

Table 1: The comparisons of the sample complexity, iteration complexity, and computational complexity for different algorithms.

the scale of $\mathbf{U}$ and $\mathbf{V}$ are the same:

$$\min_{\substack{\mathbf{U}\in\mathbb{R}^{d_1\times r}\\ \mathbf{V}\in\mathbb{R}^{d_2\times r}}} f(\mathbf{U},\mathbf{V}) := \mathcal{L}(\mathbf{U}\mathbf{V}^\top) + \frac{1}{8}\|\mathbf{U}^\top\mathbf{U} - \mathbf{V}^\top\mathbf{V}\|_F^2. \tag{3.1}$$

As discussed before in (1.4), to apply the idea of stochastic variance reduced gradient decent algorithm, we accordingly decompose the objective function $f(\mathbf{U},\mathbf{V})$ into $n$ components such that

$$\mathcal{L}(\mathbf{U}\mathbf{V}^\top) = \frac{1}{n}\sum_{i=1}^n \ell_i(\mathbf{U}\mathbf{V}^\top), \quad f(\mathbf{U},\mathbf{V}) = \frac{1}{n}\sum_{i=1}^n f_i(\mathbf{U},\mathbf{V}), \tag{3.2}$$

where for each component function, we have

$$\ell_i(\mathbf{U}\mathbf{V}^\top) = \frac{1}{2b}\sum_{j=1}^b \big(\langle \mathbf{A}_{i_j}, \mathbf{U}\mathbf{V}^\top\rangle - y_{i_j}\big)^2, \quad f_i(\mathbf{U},\mathbf{V}) = \ell_i(\mathbf{U}\mathbf{V}^\top) + \frac{1}{8}\|\mathbf{U}^\top\mathbf{U} - \mathbf{V}^\top\mathbf{V}\|_F^2. \tag{3.3}$$

Therefore, motivated by the idea of stochastic variance reduced gradient (Johnson and Zhang, 2013), we propose an accelerated stochastic gradient descent algorithm, as displayed in Algorithm 1, for the nonconvex optimization problem (3.1). The key idea of this algorithm is to reduce the variance of the stochastic gradient in each iteration and accelerate the rate of convergence.

**Algorithm 1** Stochastic Variance Reduced Gradient Descent for Matrix Sensing
1: **Input:** $\{\mathbf{A}_i, y_i\}_{i=1}^n$; step size $\eta$; number of iterations $S, m$; initial solution $(\widetilde{\mathbf{U}}^0, \widetilde{\mathbf{V}}^0)$.
2:   **for:** $s = 1, 2, \ldots S$ **do**
3:     $\widetilde{\mathbf{U}} = \widetilde{\mathbf{U}}^{s-1}, \widetilde{\mathbf{V}} = \widetilde{\mathbf{V}}^{s-1}$
4:     $\widetilde{\mathbf{G}}_{\mathbf{U}} = \nabla_{\mathbf{U}} \mathcal{L}(\widetilde{\mathbf{U}} \widetilde{\mathbf{V}}^\top)$
5:     $\widetilde{\mathbf{G}}_{\mathbf{V}} = \nabla_{\mathbf{V}} \mathcal{L}(\widetilde{\mathbf{U}} \widetilde{\mathbf{V}}^\top)$
6:     $\mathbf{U}^0 = \widetilde{\mathbf{U}}, \mathbf{V}^0 = \widetilde{\mathbf{V}}$
7:     **for:** $t = 0, 1, 2, \ldots, m-1$ **do**
8:       Randomly pick $i_t \in \{1, 2, \ldots, n\}$
9:       $\mathbf{U}^{t+1} = \mathbf{U}^t - \eta \big( \nabla_{\mathbf{U}} f_{i_t}(\mathbf{U}^t, \mathbf{V}^t) - \nabla_{\mathbf{U}} \ell_{i_t}(\widetilde{\mathbf{U}} \widetilde{\mathbf{V}}^\top) + \widetilde{\mathbf{G}}_{\mathbf{U}} \big)$
10:      $\mathbf{V}^{t+1} = \mathbf{V}^t - \eta \big( \nabla_{\mathbf{V}} f_{i_t}(\mathbf{U}^t, \mathbf{V}^t) - \nabla_{\mathbf{V}} \ell_{i_t}(\widetilde{\mathbf{U}} \widetilde{\mathbf{V}}^\top) + \widetilde{\mathbf{G}}_{\mathbf{V}} \big)$
11:     **end for**
12:     $\widetilde{\mathbf{U}}^s = \mathbf{U}^t, \widetilde{\mathbf{V}}^s = \mathbf{V}^t$ for randomly chosen $t \in \{0, \ldots, m-1\}$
13:   **end for**
14: **Output:** $(\widetilde{\mathbf{U}}^S, \widetilde{\mathbf{V}}^S)$.

Algorithm 1 provides us an efficient way to solve the matrix sensing problem using stochastic gradient descent. Compared with the original stochastic variance reduced gradient descent (Johnson and Zhang, 2013), Algorithm 1 updates $\mathbf{U}, \mathbf{V}$ simultaneously. Besides, in order to ensure the linear convergence of Algorithm 1, we need to guarantee the initial solution $(\widetilde{\mathbf{U}}, \widetilde{\mathbf{V}})$ falls into a near neighbourhood of $(\mathbf{U}^*, \mathbf{V}^*)$ (See Section 4 for a detailed argument). Thus we ultilize the projected gradient descent based initialization algorithm proposed in Tu et al. (2015), as shown in Algorithm 2. We denote the singular value decomposition of any rank-$r$ matrix $\mathbf{X} \in \mathbb{R}^{d_1 \times d_2}$ by $\text{SVD}_r(\mathbf{X})$. Suppose $\text{SVD}_r(\mathbf{X}) = [\mathbf{U}, \mathbf{\Sigma}, \mathbf{V}]$, then we let the best rank-$r$ approximation of $\mathbf{X}$ to be $\mathcal{P}_r(\mathbf{X}) = \mathbf{U}\mathbf{\Sigma}\mathbf{V}^\top$, where $\mathcal{P}_r : \mathbb{R}^{d_1 \times d_2} \to \mathbb{R}^{d_1 \times d_2}$ represents the corresponding projection operator.

**Algorithm 2** Initialization
1: **Input:** $\{\mathbf{A}_i, y_i\}_{i=1}^n$; step size $\tau$; number of iterations S.
2:   $\mathbf{X}_0 = \mathbf{0}$
3:   **for:** $s = 1, 2, 3, \ldots, S$ **do**
4:     $\mathbf{X}_s = \mathcal{P}_r[\mathbf{X}_{s-1} - \tau \nabla \mathcal{L}(\mathbf{X}_{s-1})]$
5:   **end for**
6:   $[\overline{\mathbf{U}}^0, \mathbf{\Sigma}^0, \overline{\mathbf{V}}^0] = \text{SVD}_r(\mathbf{X}_S)$
7:   $\widetilde{\mathbf{U}}^0 = \overline{\mathbf{U}}^0 (\mathbf{\Sigma}^0)^{1/2}, \widetilde{\mathbf{V}}^0 = \overline{\mathbf{V}}^0 (\mathbf{\Sigma}^0)^{1/2}$
8: **Output:** $(\widetilde{\mathbf{U}}^0, \widetilde{\mathbf{V}}^0)$.

Here in Algorithm 2, plugging in the formula of $\mathcal{L}$ in (1.3) for iteration $s$, we have

$$\nabla \mathcal{L}(\mathbf{X}_{s-1}) = \frac{1}{N} \sum_{i=1}^{N} \big( \langle \mathbf{A}_i, \mathbf{X}_{s-1} \rangle - y_i \big) \mathbf{A}_i.$$

As shown later in our theoretical analysis, in order to ensure the initial solution $(\widetilde{\mathbf{U}}^0, \widetilde{\mathbf{V}}^0)$ to be sufficiently close to the true parameter $(\mathbf{U}^*, \mathbf{V}^*)$, the initialization algorithm requires at least

$N = O(rd')$ measurements.

## 4 Main Theory

We present our main theoretical results for Algorithms 1 and 2 in this section. We first introduce the following notations to simplify our proof. Consider the singular value decomposition (SVD) of the unknown low-rank matrix $\mathbf{X}^*$ as $\mathbf{X}^* = \overline{\mathbf{U}}^* \mathbf{\Sigma}^* \overline{\mathbf{V}}^{*\top}$, where $\overline{\mathbf{U}}^* \in \mathbb{R}^{d_1 \times r}$, $\overline{\mathbf{V}}^* \in \mathbb{R}^{d_2 \times r}$ are orthonormal such that $\overline{\mathbf{U}}^{*\top} \overline{\mathbf{U}}^* = \mathbf{I}_r, \overline{\mathbf{V}}^{*\top} \overline{\mathbf{V}}^* = \mathbf{I}_r$, and $\mathbf{\Sigma}^*$ is an diagonal matrix. Let the sorted nonzero singular values of $\mathbf{X}^*$ to be $\sigma_1 \geq \sigma_2 \geq \cdots \geq \sigma_r > 0$, and the condition number of $\mathbf{X}^*$ to be $\kappa$, i.e., $\kappa = \sigma_1 / \sigma_r$. Besides, we use $\mathbf{U}^* = \overline{\mathbf{U}}^* (\mathbf{\Sigma}^*)^{1/2}$ and $\mathbf{V}^* = \overline{\mathbf{V}}^* (\mathbf{\Sigma}^*)^{1/2}$ to denote the true parameter, then we lift $\mathbf{X}^* \in \mathbb{R}^{d_1 \times d_2}$ to a positive semidefinite matrix $\mathbf{Y}^* \in \mathbb{R}^{(d_1+d_2) \times (d_1+d_2)}$, following Tu et al. (2015); Zheng and Lafferty (2016), in higher dimension as follows

$$\mathbf{Y}^* = \begin{bmatrix} \mathbf{U}^* \mathbf{U}^{*\top} & \mathbf{U}^* \mathbf{V}^{*\top} \\ \mathbf{V}^* \mathbf{U}^{*\top} & \mathbf{V}^* \mathbf{V}^{*\top} \end{bmatrix} = \mathbf{Z}^* \mathbf{Z}^{*\top},$$

where $\mathbf{Z}^*$ is defined in higher dimension as

$$\mathbf{Z}^* = \begin{bmatrix} \mathbf{U}^* \\ \mathbf{V}^* \end{bmatrix} \in \mathbb{R}^{(d_1+d_2) \times r}.$$

Noticing the symmetric factorization of $\mathbf{Y}^*$ is not unique, we define the corresponding solution set $\mathcal{Z}$ in terms of the true parameter $\mathbf{Z}^*$ as

$$\mathcal{Z} = \left\{ \mathbf{Z} \in \mathbb{R}^{(d_1+d_2) \times r} \,\Big|\, \mathbf{Z} = \mathbf{Z}^* \mathbf{R} \text{ for some } \mathbf{R} \in \mathbb{Q}_r \right\},$$

where $\mathbb{Q}_r$ represents the set of $r$-by-$r$ orthonormal matrices. Notice that for any $\mathbf{Z} \in \mathcal{Z}$, we can get $\mathbf{X}^* = \mathbf{Z}_U \mathbf{Z}_V^\top$, where $\mathbf{Z}_U$ and $\mathbf{Z}_V$ represent the top $d_1 \times r$ and bottom $d_2 \times r$ matrix of $\mathbf{Z} \in \mathbb{R}^{(d_1+d_2) \times r}$, respectively.

**Definition 4.1.** Define $d(\mathbf{Z}, \mathbf{Z}^*)$ as the distance (in terms of Frobenius norm) between $\mathbf{Z}$ and $\mathbf{Z}^*$ with respect to the optimal rotation such that

$$d(\mathbf{Z}, \mathbf{Z}^*) = \min_{\widetilde{\mathbf{Z}} \in \mathcal{Z}} \|\mathbf{Z} - \widetilde{\mathbf{Z}}\|_F = \min_{\mathbf{R} \in \mathbb{Q}_r} \|\mathbf{Z} - \mathbf{Z}^* \mathbf{R}\|_F.$$

**Definition 4.2.** Define the neighbourhood of $\mathbf{Z}^*$ with radius $R$ as

$$\mathbb{B}(R) = \left\{ \mathbf{Z} \in \mathbb{R}^{(d_1+d_2) \times r} \,\Big|\, d(\mathbf{Z}, \mathbf{Z}^*) \leq R \right\}.$$

Next, we lay out the definition of restricted isometry property, which characterizes the structure of the linear measurement operator. This restricted isometry property is essential to derive our main results regarding our proposed algorithms.

**Definition 4.3.** The linear measurement operator $\mathcal{A}_M$ is said to satisfy the restricted isometry property of order $r$ (r-RIP) with parameter $\delta_r$, if for all matrices $\mathbf{X} \in \mathbb{R}^{d_1 \times d_2}$ with $\text{rank}(\mathbf{X}) \leq r$, the

following holds

$$(1 - \delta_r)\|\mathbf{X}\|_F^2 \leq \frac{1}{M}\|\mathcal{A}_M(\mathbf{X})\|_2^2 \leq (1 + \delta_r)\|\mathbf{X}\|_F^2.$$

The restricted isometry property is commonly used in the existing literature (Recht et al., 2010; Jain et al., 2013; Chen and Wainwright, 2015; Tu et al., 2015) for the problem of linear regression, and it holds for various random ensembles, such as Gaussian ensemble $\big((\mathbf{A}_i)_{jk} \sim N(0,1)\big)$ and Rademacher ensemble $\big((\mathbf{A}_i)_{jk} \in \{-1, +1\}$ equiprobably$\big)$. For such ensembles, with the number of measurements ($M$) sufficiently large, the linear measurement operator $\mathcal{A}_M$ can satisfy the RIP with high probability. For example, for Gaussian ensemble, it is known that a $r$-RIP is satisfied with parameter $\delta_r$ if we have $M = \Omega(\delta_r^{-2} r d')$ measurements. Note that Definition 4.3 is slightly different from that in Tu et al. (2015), because we assumed that the entries of $\mathbf{A}_i$ are sampled from a standard Gaussian distribution. While in Tu et al. (2015), they consider the case that each entry of $\mathbf{A}_i$ is i.i.d. Gaussian with zero mean and variance $1/M$.

The following assumption characterizes the noise vector $\boldsymbol{\epsilon}$, which is essential to derive the statistical error rate regarding our returned estimator.

**Assumption 4.4.** Suppose that the noise vector $\boldsymbol{\epsilon}$ is bounded with respect to the number of observations for each component function, i.e., there exists a constant $\nu > 0$ such that $\|\boldsymbol{\epsilon}\|_2 \leq 2\nu\sqrt{b}$.

It is obvious that Assumption 4.4 holds for any bounded noise, and for any sub-Gaussian random noise with parameter $\nu$, it is proved in Vershynin (2010) that this assumption can hold with high probability.

Now we are ready to provide the theoretical guarantees for the stochastic variance reduced gradient descent Algorithm 1 and initialization Algorithm 2. Recall that we denote $d' = \max\{d_1, d_2\}$.

**Theorem 4.5.** Let $\mathbf{X}^* = \mathbf{U}^*\mathbf{V}^{*\top}$ be the unknown rank-$r$ matrix. Suppose the linear measurement operator $\mathcal{A}_N$ satisfies the $4r$-RIP with parameter $\delta_{4r} \in (0, 1/16)$, and for any $i \in [n]$, the linear measurement operator $\mathcal{A}_b^i$ satisfies the $4r$-RIP with parameter $\delta_{4r}' \in (0,1)$. Furthermore, assume the noise vector $\boldsymbol{\epsilon}$ satisfies Assumption 4.4. Then for any $\widetilde{\mathbf{Z}}^0 = [\widetilde{\mathbf{U}}^0; \widetilde{\mathbf{V}}^0] \in \mathbb{B}(c_2\sqrt{\sigma_r})$ with $c_2 \leq 1/4$, if the step size $\eta = c_1/\sigma_1$ and the number of iterations $m$ are properly chosen, such that the following conditions are satisfied

$$c_1 \leq \frac{1}{256(1 + \delta_{4r}')^2}, \quad \rho = 15\kappa\bigg(\frac{1}{\eta\sigma_1 m} + 384\eta\sigma_1(1 + \delta_{4r}')^2\bigg) < 1,$$

the estimator $\widetilde{\mathbf{Z}}^S = [\widetilde{\mathbf{U}}^S; \widetilde{\mathbf{V}}^S]$ from the Algorithm 1 satisfies

$$\mathbb{E}\big[d^2(\widetilde{\mathbf{Z}}^S, \mathbf{Z}^*)\big] \leq \rho^S d^2(\widetilde{\mathbf{Z}}^0, \mathbf{Z}^*) + \frac{15c_3\nu^2}{(1-\rho)\sigma_r} \cdot \bigg(\frac{rd'}{N} + \frac{\eta\sigma_1 rd'}{b}\bigg). \qquad (4.1)$$

with probability at least $1 - c_3\exp\big(-c_4 d'\big)$.

**Remark 4.6.** Theorem 4.5 implies that in order to achieve linear convergence rate, we need to set the step size $\eta$ to be sufficiently small and the inner loop iterations $m$ to be sufficiently large such that $\rho < 1$. Here we provide an example to show that this condition is absolutely achievable. As stated in the theorem, if we choose the step size $\eta = c_1'/\sigma_1$, where $c_1' = 1/(576\kappa(1+\delta_{4r})^2)$, then the contraction parameter $\rho$ will be simplified as

$$\rho = \frac{15\kappa}{\eta\sigma_1 m} + \frac{2}{3}.$$

Therefore, provided that the inner loop iterations $m$ is chosen to be $m \geq c_5\kappa^2$, we have $\rho \leq 5/6 < 1$.

**Remark 4.7.** The right hand side of (4.1) consists of two terms, the first term corresponds to the optimization error and the second term corresponds to the statistical error. More specifically, in the noisy case, with a appropriate number of inner loop iterations $m$, after $O\big(\log(N/(rd'))\big)$ number of outer loop iterations, the estimator returned by our algorithm achieves $O(\sqrt{rd'/N})$ statistical error since $b \asymp N$, which matches the optimal minimax lower bound for matrix sensing problem (Negahban and Wainwright, 2011). While in the noiseless case, since $\nu$ is 0, the second term in (4.1) becomes 0. In order to satisfy 4$r$-RIP for linear measurement operators, we need the sample size $N = O(rd')$, which obtains the optimal sample complexity that is required for matrix sensing problem (Recht et al., 2010; Tu et al., 2015; Wang et al., 2016). Most importantly, Theorem 4.5 implies that, to achieve $\epsilon$ accuracy for optimization error, our algorithm requires $O\big(\log(1/\epsilon)\big)$ outer iterations. Since for each outer iteration, we need to calculate $m = O(\kappa^2)$ stochastic variance reduced gradients and one full gradient, the overall computational complexity for our algorithm to reach $\epsilon$ accuracy is

$$O\bigg((Nd'^2 + \kappa^2 bd'^2)\log\bigg(\frac{1}{\epsilon}\bigg)\bigg).$$

However, for standard full gradient descent algorithms, the overall computational complexity for the state-of-the-art algorithms for noiseless (Tu et al., 2015) and noisy case (Wang et al., 2016) to reach $\epsilon$ accuracy is $O\big(N\kappa d'^2 \log(1/\epsilon)\big)$. Therefore, if we have $\kappa \leq n$, our method is more efficient than the state-of-the-art gradient descent methods. The more complete comparison with existing algorithms is summarized in Table 1.

Finally, we present the theoretical guarantees for the initialization Algorithm 2, and we refer to Wang et al. (2016) for a detailed proof of Algorithm 2.

**Theorem 4.8.** Let $\mathbf{X}^* = \mathbf{U}^*\mathbf{V}^{*\top}$ be the unknown rank-$r$ matrix. Consider $\widetilde{\mathbf{X}}^0 = \widetilde{\mathbf{U}}^0\widetilde{\mathbf{V}}^{0\top}$, where $\widetilde{\mathbf{U}}^0, \widetilde{\mathbf{V}}^0$ are produced in the initialization Algorithm 2. Suppose the linear measurement operator $\mathcal{A}_N$ satisfies 4$r$-RIP with parameter $\delta_{4r} \in (0, 1/2)$ and the noise vector $\boldsymbol{\epsilon}$ satisfies Assumption 4.4, then there exist constants $c_1, c_2, c_3$ such that

$$\|\widetilde{\mathbf{X}}^0 - \mathbf{X}^*\|_F \leq \rho^S \|\mathbf{X}^*\|_F + \frac{c_1\nu}{1-\rho} \cdot \sqrt{\frac{rd'}{N}}, \quad (4.2)$$

holds with probability at least $1 - c_2\exp(-c_3 d')$, where $\rho = 2\delta_{4r} \in (0, 1)$ is the contraction parameter.

**Remark 4.9.** The right hand side of (4.2) consists of two terms, the first term is the optimization error and the second term corresponds to the statistical error. To satisfy the initial constraint $\widetilde{\mathbf{Z}}^0 \in \mathbb{B}(\sqrt{\sigma_r}/4)$ in Theorem 4.5, it suffices to guarantee $\widetilde{\mathbf{X}}^0$ is close enough to the unknown rank-$r$ matrix $\mathbf{X}^*$, i.e., $\|\widetilde{\mathbf{X}}^0 - \mathbf{X}^*\|_F \leq \sigma_r/2$. In fact, according to lemma B.2, since $\|\widetilde{\mathbf{X}}^0 - \mathbf{X}^*\|_2 \leq \|\widetilde{\mathbf{X}}^0 - \mathbf{X}^*\|_F \leq \sigma_r/2$, we have

$$d^2(\widetilde{\mathbf{Z}}^0, \mathbf{Z}^*) \leq \frac{\sqrt{2}-1}{2} \cdot \frac{\|\widetilde{\mathbf{X}}^0 - \mathbf{X}^*\|_F^2}{\sigma_r} \leq \frac{\sigma_r}{16}.$$

Besides, as for the statistical error term in (4.2), we can always set the number of measurements $N \geq c \cdot rd'$, where $c$ is a constant large enough such that the statistical error is $O(\sigma_r)$. Therefore, in order to meet the initial ball requirement of Theorem 4.5, it is sufficient to perform $S' = c' \log(\sigma_r/\|\mathbf{X}^*\|_F)/(\log \rho) = O(1)$ iterations in Algorithm 2, where $c'$ is a constant.

## 5 Proofs of the Main Theory

In this section, we lay out the theoretical proofs of our main results. We first introduce the following notations for simplicity. For any $\mathbf{Z} \in \mathbb{R}^{(d_1+d_2) \times r}$, recall that $\mathbf{Z} = [\mathbf{U}; \mathbf{V}]$, where $\mathbf{U} \in \mathbb{R}^{d_1 \times r}, \mathbf{V} \in \mathbb{R}^{d_2 \times r}$. According to (3.1), the objective function in terms of $\mathbf{Z}$ we intend to minimize is as follows

$$\widetilde{f}(\mathbf{Z}) = f(\mathbf{U}, \mathbf{V}) = \mathcal{L}(\mathbf{U}\mathbf{V}^\top) + \frac{1}{8}\|\mathbf{U}^\top\mathbf{U} - \mathbf{V}^\top\mathbf{V}\|_F^2 = \frac{1}{2N}\sum_{i=1}^N \left(\langle \mathbf{A}_i, \mathbf{U}\mathbf{V}^\top \rangle - y_i\right)^2 + \frac{1}{8}\|\mathbf{U}^\top\mathbf{U} - \mathbf{V}^\top\mathbf{V}\|_F^2,$$

Therefore, we have the gradient of $\widetilde{f}(\mathbf{Z})$ as

$$\nabla \widetilde{f}(\mathbf{Z}) = \begin{bmatrix} \nabla_\mathbf{U}\mathcal{L}(\mathbf{U}\mathbf{V}^\top) + \frac{1}{2}\mathbf{U}(\mathbf{U}^\top\mathbf{U} - \mathbf{V}^\top\mathbf{V}) \\ \nabla_\mathbf{V}\mathcal{L}(\mathbf{U}\mathbf{V}^\top) + \frac{1}{2}\mathbf{V}(\mathbf{U}^\top\mathbf{U} - \mathbf{V}^\top\mathbf{V}) \end{bmatrix}, \quad (5.1)$$

where

$$\nabla_\mathbf{U}\mathcal{L}(\mathbf{U}\mathbf{V}^\top) = \frac{1}{N}\sum_{i=1}^N (\langle \mathbf{A}_i, \mathbf{U}\mathbf{V}^\top \rangle - y_i)\mathbf{A}_i\mathbf{V}, \quad \nabla_\mathbf{V}\mathcal{L}(\mathbf{U}\mathbf{V}^\top) = \frac{1}{N}\sum_{i=1}^N (\langle \mathbf{A}_i, \mathbf{U}\mathbf{V}^\top \rangle - y_i)\mathbf{A}_i^\top\mathbf{U}.$$

Besides, for each component function in (3.2), we denote $\widetilde{\ell}_i(\mathbf{Z}) = \ell_i(\mathbf{U}\mathbf{V}^\top)$ and $\widetilde{f}_i(\mathbf{Z}) = f_i(\mathbf{U}, \mathbf{V})$. Obviously, we have $\widetilde{f}(\mathbf{Z}) = \sum_{i=1}^n \widetilde{f}_i(\mathbf{Z})/n$, where we have

$$\nabla \widetilde{f}_i(\mathbf{Z}) = \begin{bmatrix} \nabla_\mathbf{U} f_i(\mathbf{U}\mathbf{V}^\top) \\ \nabla_\mathbf{V} f_i(\mathbf{U}\mathbf{V}^\top) \end{bmatrix} = \begin{bmatrix} \nabla_\mathbf{U}\ell_i(\mathbf{U}\mathbf{V}^\top) + \frac{1}{2}\mathbf{U}(\mathbf{U}^\top\mathbf{U} - \mathbf{V}^\top\mathbf{V}) \\ \nabla_\mathbf{V}\ell_i(\mathbf{U}\mathbf{V}^\top) + \frac{1}{2}\mathbf{V}(\mathbf{V}^\top\mathbf{V} - \mathbf{U}^\top\mathbf{U}) \end{bmatrix}. \quad (5.2)$$

### 5.1 Proof of Theorem 4.5

In order to prove Theorem 4.5, we need to make use of the following lemmas. Under the RIP condition, Lemma 5.1 shows that the regularized loss function $\widetilde{f}$ satisfies a local curvature condition, and Lemma 5.2 shows that each component function $\widetilde{f}_i$ satisfies a local smoothness condition. We present their proofs in Sections A.1 and A.2, repectively.

**Lemma 5.1** (Local Curvature Condition). Suppose the transformation operator $\mathcal{A}_N$ satisfies the $4r$-RIP condition with parameter $\delta_{4r} \in (0, 1/16)$. Denote $\mathbf{R} = \operatorname{argmin}_{\widetilde{\mathbf{R}} \in \mathbb{Q}_r} \|\mathbf{Z} - \mathbf{Z}^*\widetilde{\mathbf{R}}\|_F$ by the optimal rotation with respect to $\mathbf{Z}$, and let $\mathbf{H} = \mathbf{Z} - \mathbf{Z}^*\mathbf{R}$, then we have following holds

$$\langle \nabla \widetilde{f}(\mathbf{Z}), \mathbf{H}\rangle \geq \frac{\sigma_r}{10}\|\mathbf{H}\|_F^2 + \frac{1}{8}\|\mathbf{X} - \mathbf{X}^*\|_F^2 + \frac{1}{16}\|\widetilde{\mathbf{Z}}^\top \mathbf{Z}\|_F^2 - \frac{1}{3}\|\mathbf{H}\|_F^4 - 32r\left\|\frac{1}{N}\sum_{i=1}^N \epsilon_i \mathbf{A}_i\right\|_2^2.$$

**Lemma 5.2** (Local Smoothness Condition). Let $\mathbf{Z} = [\mathbf{U}; \mathbf{V}]$ and the $i$-th component function $\widetilde{f}_i(\mathbf{Z}) = f_i(\mathbf{U}, \mathbf{V}) = \ell_i(\mathbf{U}\mathbf{V}^\top) + \|\mathbf{U}^\top\mathbf{U} - \mathbf{V}^\top\mathbf{V}\|_F^2/8$. Suppose the corresponding transformation operator $\mathcal{A}_b^i$ satisfies $4r$-RIP condition with parameter $\delta'_{4r} \in (0,1)$, then we have

$$\left\|\nabla \widetilde{f}_i(\mathbf{Z})\right\|_F^2 \leq \left(8(1+\delta'_{4r})^2 \cdot \|\mathbf{X} - \mathbf{X}^*\|_F^2 + \|\mathbf{U}^\top\mathbf{U} - \mathbf{V}^\top\mathbf{V}\|_F^2\right)\cdot \|\mathbf{Z}\|_2^2 + 8r\left\|\frac{1}{b}\sum_{j\in[b]} \epsilon_{i_j}\mathbf{A}_{i_j}\right\|_2^2 \cdot \|\mathbf{Z}\|_2^2.$$

Furthermore, if we let $\widetilde{\ell}_i(\mathbf{Z}) = \ell_i(\mathbf{U}\mathbf{V}^\top)$, by the same techniques, we have

$$\left\|\nabla \widetilde{\ell}_i(\mathbf{Z})\right\|_F^2 \leq 8(1+\delta'_{4r})^2 \cdot \|\mathbf{X} - \mathbf{X}^*\|_F^2 \cdot \|\mathbf{Z}\|_2^2 + 8r\left\|\frac{1}{b}\sum_{j\in[b]} \epsilon_{i_j}\mathbf{A}_{i_j}\right\|_2^2 \cdot \|\mathbf{Z}\|_2^2.$$

The following lemma upper bounds the statistical error term returned by our algorithm, which was used in Negahban and Wainwright (2011), as long as the noise vector satisfies Assumption 4.4.

**Lemma 5.3.** Consider the linear measurement operator $\mathcal{A}_M$ with element $\{\mathbf{A}_i\}_{i=1}^M$ sampled from $\mathbf{\Sigma}$-ensemble. In addition, suppose the noise vector $\boldsymbol{\epsilon}$ satisfies Assumption 4.4 such that $\|\boldsymbol{\epsilon}\|_2 \leq 2\nu\sqrt{M}$. Then there exist constants $C, C_1$ and $C_2$ such that the following holds with probability at least $1 - C_1\exp(-C_2 d')$

$$\left\|\frac{1}{M}\sum_{i=1}^M \epsilon_i \mathbf{A}_i\right\|_2 \leq C\nu\sqrt{\frac{d'}{M}}.$$

Now we are ready to prove Theorem 4.5.

*Proof of Theorem 4.5.* Denote $\mathbf{Z}^t = [\mathbf{U}^t; \mathbf{V}^t]$. According to Algorithm 1, the stochastic gradient descent based update can be rewritten as

$$\mathbf{Z}^{t+1} = \mathbf{Z}^t - \eta\big(\nabla \widetilde{f}_{i_t}(\mathbf{Z}^t) - \nabla \widetilde{\ell}_{i_t}(\widetilde{\mathbf{Z}}) + \widetilde{\mathbf{G}}\big) = \mathbf{Z}^t - \eta\mathbf{G}^t,$$

where $\mathbf{G}^t = \nabla \widetilde{f}_{i_t}(\mathbf{Z}^t) - \nabla \widetilde{\ell}_{i_t}(\widetilde{\mathbf{Z}}) + \widetilde{\mathbf{G}}$. For simplicity, we define $\mathbf{R}^t = \operatorname{argmin}_{\mathbf{R}\in\mathbb{Q}_r}\|\mathbf{Z}^t - \mathbf{Z}^*\mathbf{R}\|_F$ as the optimal rotation with respect to $\mathbf{Z}^t$, and denote $\mathbf{H}^t = \mathbf{Z} - \mathbf{Z}^*\mathbf{R}^t$, which implies that $d^2(\mathbf{Z}^t, \mathbf{Z}^*) = \|\mathbf{H}^t\|_F^2$. Similarly, we use $\widetilde{\mathbf{R}}$ to denote the optimal rotation with respect to $\widetilde{\mathbf{Z}}$, such that $\widetilde{\mathbf{R}} = \operatorname{argmin}_{\mathbf{R}\in\mathbb{Q}_r}\|\widetilde{\mathbf{Z}} - \mathbf{Z}^*\mathbf{R}\|_F$. By induction, we assume $\widetilde{\mathbf{Z}} \in \mathbb{B}(\sqrt{\sigma_r}/4)$, and for any $t \geq 0$, we assume $\mathbf{Z}^t \in \mathbb{B}(\sqrt{\sigma_r}/4)$. Note that $\widetilde{\mathbf{G}} = \sum_{i=1}^n \nabla \ell_i(\widetilde{\mathbf{Z}})/n$, thus by taking expectation of $\mathbf{H}^{t+1}$ over $i_t$ conditioned on $\mathbf{Z}^t$, we have

$$\begin{aligned}
\mathbb{E}\|\mathbf{H}^{t+1}\|_F^2 &\leq \mathbb{E}\|\mathbf{Z}^t - \eta \mathbf{G}^t - \mathbf{Z}^*\mathbf{R}^t\|_F^2 \\
&= \|\mathbf{H}^t\|_F^2 + \eta^2 \mathbb{E}\|\mathbf{G}^t\|_F^2 - 2\eta \mathbb{E}\langle \mathbf{H}^t, \mathbf{G}^t \rangle \\
&= \|\mathbf{H}^t\|_F^2 + \eta^2 \mathbb{E}\|\mathbf{G}^t\|_F^2 - 2\eta \langle \nabla \widetilde{f}(\mathbf{Z}^t), \mathbf{H}^t \rangle, \quad (5.3)
\end{aligned}$$

where the first inequality is due to the definition of $\mathbf{H}^t$, and the last equality holds because conditioned on $\mathbf{Z}^t$, $\mathbb{E}\langle \mathbf{H}^t, \mathbf{G}^t \rangle = \langle \mathbf{H}^t, \mathbb{E}\mathbf{G}^t \rangle = \langle \mathbf{H}^t, \nabla \widetilde{f}(\mathbf{Z}^t) \rangle$. Note that we assume $\delta_{4r} \in (0, 1/16)$. Then according to Lemma 5.1, there exist constants $C_1, C_2$ and $C_3$ such that with probability at least $1 - C_1 \exp(-C_2 d')$, we have

$$\begin{aligned}
\langle \nabla \widetilde{f}(\mathbf{Z}^t), \mathbf{H}^t \rangle &\geq \frac{\sigma_r}{10} \|\mathbf{H}^t\|_F^2 + \frac{1}{8}\|\mathbf{X}^t - \mathbf{X}^*\|_F^2 + \frac{1}{16}\|\mathbf{U}^{t\top}\mathbf{U}^t - \mathbf{V}^{t\top}\mathbf{V}^t\|_F^2 - \frac{1}{3}\|\mathbf{H}^t\|_F^4 - 32r\left\|\frac{1}{N}\sum_{i=1}^N \epsilon_i \mathbf{A}_i\right\|_2^2 \\
&\geq \frac{\sigma_r}{10} \|\mathbf{H}^t\|_F^2 + \frac{1}{8}\|\mathbf{X}^t - \mathbf{X}^*\|_F^2 + \frac{1}{16}\|\mathbf{U}^{t\top}\mathbf{U}^t - \mathbf{V}^{t\top}\mathbf{V}^t\|_F^2 - \frac{1}{3}\|\mathbf{H}^t\|_F^4 - C_3 \nu^2 \frac{rd'}{N}, \quad (5.4)
\end{aligned}$$

where the last inequality follows from Lemma 5.3. Thus, it is sufficient to upper bound the term $\mathbb{E}\|\mathbf{G}^t\|_F^2$. Plugging in the formula of $\mathbf{G}^t$, we have

$$\begin{aligned}
\mathbb{E}\|\mathbf{G}^t\|_F^2 &= \mathbb{E}\|\nabla \widetilde{f}_{i_t}(\mathbf{Z}^t) - \nabla \widetilde{\ell}_{i_t}(\widetilde{\mathbf{Z}}) + \widetilde{\mathbf{G}}\|_F^2 \\
&\leq 2\mathbb{E}\|\nabla \widetilde{f}_{i_t}(\mathbf{Z}^t)\|_F^2 + 2\mathbb{E}\|\nabla \widetilde{\ell}_{i_t}(\widetilde{\mathbf{Z}}) - \mathbb{E}[\nabla \widetilde{\ell}_{i_t}(\widetilde{\mathbf{Z}})]\|_F^2 \\
&\leq 2\mathbb{E}\|\nabla \widetilde{f}_{i_t}(\mathbf{Z}^t)\|_F^2 + 2\mathbb{E}\|\nabla \widetilde{\ell}_{i_t}(\widetilde{\mathbf{Z}})\|_F^2,
\end{aligned}$$

where the expectation is taken with respect to $i_t$. The first inequality holds because $\widetilde{\mathbf{G}} = \mathbb{E}[\nabla \widetilde{\ell}_{i_t}(\widetilde{\mathbf{Z}})]$ and $\|\mathbf{A} + \mathbf{B}\|_F^2 \leq 2\|\mathbf{A}\|_F^2 + 2\|\mathbf{B}\|_F^2$, while the second inequality holds because $\mathbb{E}\|\boldsymbol{\xi} - \mathbb{E}\boldsymbol{\xi}\|_2^2 \leq \mathbb{E}\|\boldsymbol{\xi}\|_2^2$ for any random vector $\boldsymbol{\xi}$. Note that for any $\mathbf{Z} \in \mathbb{B}(\sqrt{\sigma_r}/4)$, denote $\mathbf{R}$ as the optimal rotation with respect to $\mathbf{Z}$, we have $\|\mathbf{Z}\|_2 \leq \|\mathbf{Z}^*\|_2 + \|\mathbf{Z} - \mathbf{Z}^*\mathbf{R}\|_2 \leq 2\sqrt{\sigma_1}$. Therefore, according to Lemma 5.2, we have

$$\begin{aligned}
\mathbb{E}\|\mathbf{G}^t\|_F^2 &\leq \frac{2}{n}\sum_{i=1}^n \|\nabla \widetilde{f}_i(\mathbf{Z}^t)\|_F^2 + \frac{2}{n}\sum_{i=1}^n \|\nabla \widetilde{\ell}_i(\widetilde{\mathbf{Z}})\|_F^2 \\
&\leq 16(1+\delta_{4r}')^2 \big(\|\mathbf{X}^t - \mathbf{X}^*\|_F^2 \cdot \|\mathbf{Z}^t\|_2^2 + \|\widetilde{\mathbf{X}} - \mathbf{X}^*\|_F^2 \cdot \|\widetilde{\mathbf{Z}}\|_2^2\big) \\
&\quad + 2\|\mathbf{U}^{t\top}\mathbf{U}^t - \mathbf{V}^{t\top}\mathbf{V}^t\|_F^2 \cdot \|\mathbf{Z}^t\|_2^2 + \frac{16r}{n}\sum_{i=1}^n \left\|\frac{1}{b}\sum_{j\in[b]} \epsilon_{i_j} \mathbf{A}_{i_j}\right\|_2^2 \cdot (\|\mathbf{Z}^t\|_F^2 + \|\widetilde{\mathbf{Z}}\|_F^2) \\
&\leq 64(1+\delta_{4r}')^2 \sigma_1 \big(\|\mathbf{X}^t - \mathbf{X}^*\|_F^2 + \|\widetilde{\mathbf{X}} - \mathbf{X}^*\|_F^2\big) \\
&\quad + 8\sigma_1 \|\mathbf{U}^{t\top}\mathbf{U}^t - \mathbf{V}^{t\top}\mathbf{V}^t\|_F^2 + \frac{128 r \sigma_1}{n}\sum_{i=1}^n \left\|\frac{1}{b}\sum_{j\in[b]} \epsilon_{i_j} \mathbf{A}_{i_j}\right\|_2^2, \quad (5.5)
\end{aligned}$$

where the second inequality follows from Lemma 5.2, and the last inequality holds because $\|\widetilde{\mathbf{Z}}\|_2 \leq 2\sqrt{\sigma_1}$ and $\|\mathbf{Z}^t\|_2 \leq 2\sqrt{\sigma_1}$. Furthermore, according to Lemma 5.3 and union bound, there exist

constants $C_1', C_2'$ and $C_3'$, such that with probability at least $1 - C_1' n \exp(-C_2' d')$, we have

$$\frac{1}{n} \sum_{i=1}^{n} \left\| \frac{1}{b} \sum_{j \in [b]} \epsilon_{i_j} \mathbf{A}_{i_j} \right\|_2^2 \leq C_3' \frac{\nu^2 d'}{b}. \tag{5.6}$$

Thus plugging (5.6) into (5.5), we obtain the upper bound of $\mathbb{E}\|\mathbf{G}^t\|_F^2$

$$\mathbb{E}\|\mathbf{G}^t\|_F^2 \leq 64(1+\delta_{4r}')^2 \sigma_1 \big(\|\mathbf{X}^t - \mathbf{X}^*\|_F^2 + \|\widetilde{\mathbf{X}} - \mathbf{X}^*\|_F^2\big) + 8\sigma_1 \|\mathbf{U}^{t\top}\mathbf{U}^t - \mathbf{V}^{t\top}\mathbf{V}^t\|_F^2 + 128 C_3' \nu^2 \sigma_1 \frac{rd'}{b}. \tag{5.7}$$

Note that we assume $\eta = c_1/\sigma_1$, where $c_1 \leq 1/\big(256(1+\delta_{4r}')^2\big)$. Therefore, combining (5.4) and (5.7), we have

$$\eta^2 \mathbb{E}\|\mathbf{G}^t\|_F^2 - 2\eta \langle \nabla \widetilde{f}(\mathbf{Z}^t), \mathbf{H}^t \rangle \leq -\frac{1}{5}\eta \sigma_r \|\mathbf{H}^t\|_F^2 + 64(1+\delta_{4r}')^2 \sigma_1 \eta^2 \|\widetilde{\mathbf{X}} - \mathbf{X}^*\|_F^2 + \frac{2}{3}\eta \|\mathbf{H}^t\|_F^4 \\ + 2C_3 \eta \nu^2 \frac{rd'}{N} + 128 C_3' \eta^2 \nu^2 \sigma_1 \frac{rd'}{b}, \tag{5.8}$$

holds with probability at least $1 - Cn\exp(-C'd')$. Besides, note that under our assumption, we have $\|\mathbf{H}^t\|_F^2 \leq c_2^2 \sigma_r$, where $c_2^2 \leq 1/5$. Thus by plugging (5.8) into (5.3), we have

$$\mathbb{E}\|\mathbf{H}^{t+1}\|_F^2 \leq \left(1 - \frac{\eta\sigma_r}{15}\right) \|\mathbf{H}^t\|_F^2 + 64(1+\delta_{4r}')^2 \sigma_1 \eta^2 \|\widetilde{\mathbf{X}} - \mathbf{X}^*\|_F^2 + c_3 \eta \nu^2 \left(\frac{rd'}{N} + \frac{\eta \sigma_1 rd'}{b}\right), \tag{5.9}$$

where $c_3 = \max\{2C_3, 128C_3'\}$. Finally, consider a fixed stage of $s$, such that $\widetilde{\mathbf{Z}} = \widetilde{\mathbf{Z}}^{s-1}$ and $\widetilde{\mathbf{X}} = \widetilde{\mathbf{X}}^{s-1}$, accordingly. Note that according to Algorithm 1, $\widetilde{\mathbf{Z}}^s$ is randomly selected after all of the updates are completed. By summing the previous inequality (5.9) over $t \in \{0, 1, \cdots, m-1\}$, and taking expectation with respect to all the history, we obtain

$$\mathbb{E}\|\mathbf{H}^m\|_F^2 - \mathbb{E}\|\mathbf{H}^0\|_F^2 \leq -\frac{\eta\sigma_r}{15} \sum_{t=0}^{m-1} \mathbb{E}\|\mathbf{H}^t\|_F^2 + 64(1+\delta_{4r}')^2 \sigma_1 \eta^2 m \mathbb{E}\|\widetilde{\mathbf{X}}^{s-1} - \mathbf{X}^*\|_F^2 \\ + c_3 \eta m \nu^2 \left(\frac{rd'}{N} + \frac{\eta \sigma_1 rd'}{b}\right).$$

According to the choice of $\widetilde{\mathbf{U}}^s$ and $\widetilde{\mathbf{V}}^s$ in Algorithm 1, we have

$$\mathbb{E}\|\widetilde{\mathbf{H}}^s\|_F^2 = \frac{1}{m} \sum_{t=0}^{m-1} \mathbb{E}\|\mathbf{H}^t\|_F^2,$$

where $\widetilde{\mathbf{H}}^s = \widetilde{\mathbf{Z}}^s - \mathbf{Z}^* \widetilde{\mathbf{R}}^s$, and $\widetilde{\mathbf{R}}^s = \operatorname{argmin}_{\widehat{\mathbf{R}} \in \mathbb{Q}_r} \|\widetilde{\mathbf{Z}}^s - \mathbf{Z}^* \widehat{\mathbf{R}}\|_F$. Note that $\mathbf{H}^0 = \widetilde{\mathbf{H}}^{s-1}$, we further

obtain

$$\mathbb{E}\|\mathbf{H}^m\|_F^2 - \mathbb{E}\|\widetilde{\mathbf{H}}^{s-1}\|_F^2 \leq -\frac{\eta m \sigma_r}{15}\mathbb{E}\|\widetilde{\mathbf{H}}^s\|_F^2 + 64(1+\delta'_{4r})^2\sigma_1\eta^2 m\mathbb{E}\|\widetilde{\mathbf{X}}^{s-1} - \mathbf{X}^*\|_F^2$$
$$+ c_3\eta m\nu^2\left(\frac{rd'}{N} + \frac{\eta\sigma_1 rd'}{b}\right).$$

Finally, according to Lemma B.4, for any matrix $\mathbf{Z} \in \mathbb{R}^{d_1 \times d_2}$ such that $\mathbf{Z} = [\mathbf{U}; \mathbf{V}]$ with $\mathbf{X} = \mathbf{U}\mathbf{V}^\top$, we have

$$\|\mathbf{X} - \mathbf{X}^*\|_F^2 \leq \frac{1}{2}\|\mathbf{Z}\mathbf{Z}^\top - \mathbf{Z}^*\mathbf{Z}^{*\top}\|_F^2 \leq 3\ \|\mathbf{Z}^*\|_2^2 \cdot d^2(\mathbf{Z}, \mathbf{Z}^*) = 6\sigma_1 \cdot d^2(\mathbf{Z}, \mathbf{Z}^*).$$

Therefore, we obtain

$$\frac{\eta m \sigma_r}{15}\mathbb{E}\|\widetilde{\mathbf{H}}^s\|_F^2 \leq (384\eta^2\sigma_1^2(1+\delta'_{4r})^2 m + 1) \cdot \mathbb{E}\|\widetilde{\mathbf{H}}^{s-1}\|_F^2 + c_3\eta m\nu^2\left(\frac{rd'}{N} + \frac{\eta\sigma_1 rd'}{b}\right),$$

which implies that the contraction parameter $\rho = 15\kappa\big(1/(\eta\sigma_1 m) + 384\eta\sigma_1(1+\delta'_{4r})^2\big)$. Note that we can always choose constant $c_1$ to be small enough and number of iterations $m$ to be large enough, such that $\rho \in (0,1)$. Thus, we have

$$\mathbb{E}\|\widetilde{\mathbf{H}}^s\|_F^2 \leq \rho\mathbb{E}\|\widetilde{\mathbf{H}}^{s-1}\|_F^2 + \frac{15c_3\nu^2}{\sigma_r} \cdot \left(\frac{rd'}{N} + \frac{\eta\sigma_1 rd'}{b}\right),$$

which completes the proof. □

## 6 Numerical Experiments

We use synthetic data to further study the empirical performance of our algorithm in this section.

We investigate the convergence rate of our proposed stochastic variance reduced gradient descent algorithm, and compare it with the state-of-the-art gradient descent algorithm (Tu et al., 2015). In addition, we evaluate the sample complexity, for the noiseless case, that is required by both algorithms to recover unknown low-rank matrices, and the statistical error of our algorithm for the noisy case. Both algorithms use the same initialization algorithm in Algorithm 2. Note that for the proposed stochastic variance reduced method, all results are based on the optimal iteration number $m$ and batch size $b$, which are chosen by cross validation, and averaged over 30 simulations. More specifically, we consider following settings for the unknown matrix $\mathbf{X}^*$: (i) $d_1 = 50, d_2 = 30, r = 3$; (ii) $d_1 = 50, d_2 = 30, r = 5$; (iii) $d_1 = 70, d_2 = 30, r = 3$; and (iv) $d_1 = 70, d_2 = 30, r = 5$. In all these settings, we first obtain $\mathbf{X}^* = \mathbf{U}^*\mathbf{V}^{*\top}$, where $\mathbf{U}^* \in \mathbb{R}^{d_1 \times r}, \mathbf{V}^* \in \mathbb{R}^{d_2 \times r}$ are randomly generated. Then, we generate linear measurements through the observation model $y_i = \langle \mathbf{A}_i, \mathbf{X}^*\rangle + \epsilon_i$, where each observation matrix $\mathbf{A}_i$ has i.i.d. standard Gaussian elements. In addition, we consider both (1) noisy case: each noise follows i.i.d. Gaussian distribution with zero mean and standard deviation $\sigma = 0.5$ and (2) noiseless case.

To demonstrate the rate of convergence, we report the logarithm of the squared relative error $\|\widehat{\mathbf{X}} - \mathbf{X}^*\|_F^2/\|\mathbf{X}^*\|_F^2$ versus number of effective data passes. In the noiseless case, Figure 1(a) and

1(c) show the linear rate of convergence of our algorithm, which is consistent with the convergence results of our proposed algorithm. Most importantly, it is obvious that our proposed algorithm outperforms the state-of-the-art gradient descent based algorithm in estimation error after the same number of effective data passes, which corroborates the effectiveness of our method. For other settings, we get results with similar patterns, hence we leave them out to save space.

To demonstrate the sample complexity, we report the empirical probability of exact recovery with respect to different sample size. Based on the estimator $\widehat{\mathbf{X}}$ returned by our algorithm given $N$ observations, a trial is said to be exact recovery, if we have the relative error $\|\widehat{\mathbf{X}} - \mathbf{X}^*\|_F^2/\|\mathbf{X}^*\|_F$ that is less than or equal to $10^{-3}$. The recovery probability results under setting (i) with different methods are shown in Figure 1(b). We can see that there is a phase transition around $N = 3rd'$, which confirms the optimal sample complexity $N = O(rd')$. Besides, for other settings, we get results with similar patterns, hence we leave them out to save space. Figure 1(d) shows, in the noisy case, the relationship between estimation errors and $N/(rd')$, which is consistent with our theory.

# 7 Conclusions

We proposed a stochastic variance reduced gradient descent algorithm for nonconvex low-rank matrix estimation from linear measurements under both noise and noiseless cases. We show that the proposed algorithms enjoys a linear rate of convergence to the unknown true parameter up to the minimax statistical error rate, and achieves lower computational complexity compared with the state-of-the-art approaches. Thorough experiments on synthetic datasets confirm the effectiveness of our proposed algorithms.

# A Proofs of Technical Lemmas

## A.1 Proof of Lemma 5.1

We need to make use of the following lemma to prove the local curvature condition. Denote $\widetilde{\mathbf{Z}} \in \mathbb{R}^{(d_1+d_2) \times r}$ as $\widetilde{\mathbf{Z}} = [\mathbf{U}; -\mathbf{V}]$ in the following discussions. Then the regularization term $\|\mathbf{U}^\top\mathbf{U} - \mathbf{V}^\top\mathbf{V}\|_F^2$ can be expressed as $\|\widetilde{\mathbf{Z}}^\top\mathbf{Z}\|_F^2$, and its gradient with respect to $\mathbf{Z}$ will be $4\widetilde{\mathbf{Z}}\widetilde{\mathbf{Z}}^\top\mathbf{Z}$ accordingly. Lemma A.1 shows that the regularization term satisfies a similar local curvature condition. We refer to Wang et al. (2016) for a detailed proof.

**Lemma A.1** (Local Curvature Condition for Regularization Term). *Consider $\mathbf{Z}, \mathbf{Z}^* \in \mathbb{R}^{(d_1+d_2) \times r}$. Let the optimal rotation with respect to $\mathbf{Z}$ to be $\mathbf{R} = \mathrm{argmin}_{\widetilde{\mathbf{R}} \in \mathbb{Q}_r} \|\mathbf{Z} - \mathbf{Z}^*\widetilde{\mathbf{R}}\|_F$, and let $\mathbf{H} = \mathbf{Z} - \mathbf{Z}^*\mathbf{R}$. For the gradient of the regularizer $\|\widetilde{\mathbf{Z}}^\top\mathbf{Z}\|_F^2$, we have the following holds*

$$\langle \widetilde{\mathbf{Z}}\widetilde{\mathbf{Z}}^\top\mathbf{Z}, \mathbf{H} \rangle \geq \frac{1}{2}\|\widetilde{\mathbf{Z}}^\top\mathbf{Z}\|_F^2 - \frac{1}{2}\|\widetilde{\mathbf{Z}}^\top\mathbf{Z}\|_F \cdot \|\mathbf{H}\|_F^2.$$

Now, we are ready to prove Lemma 5.1.

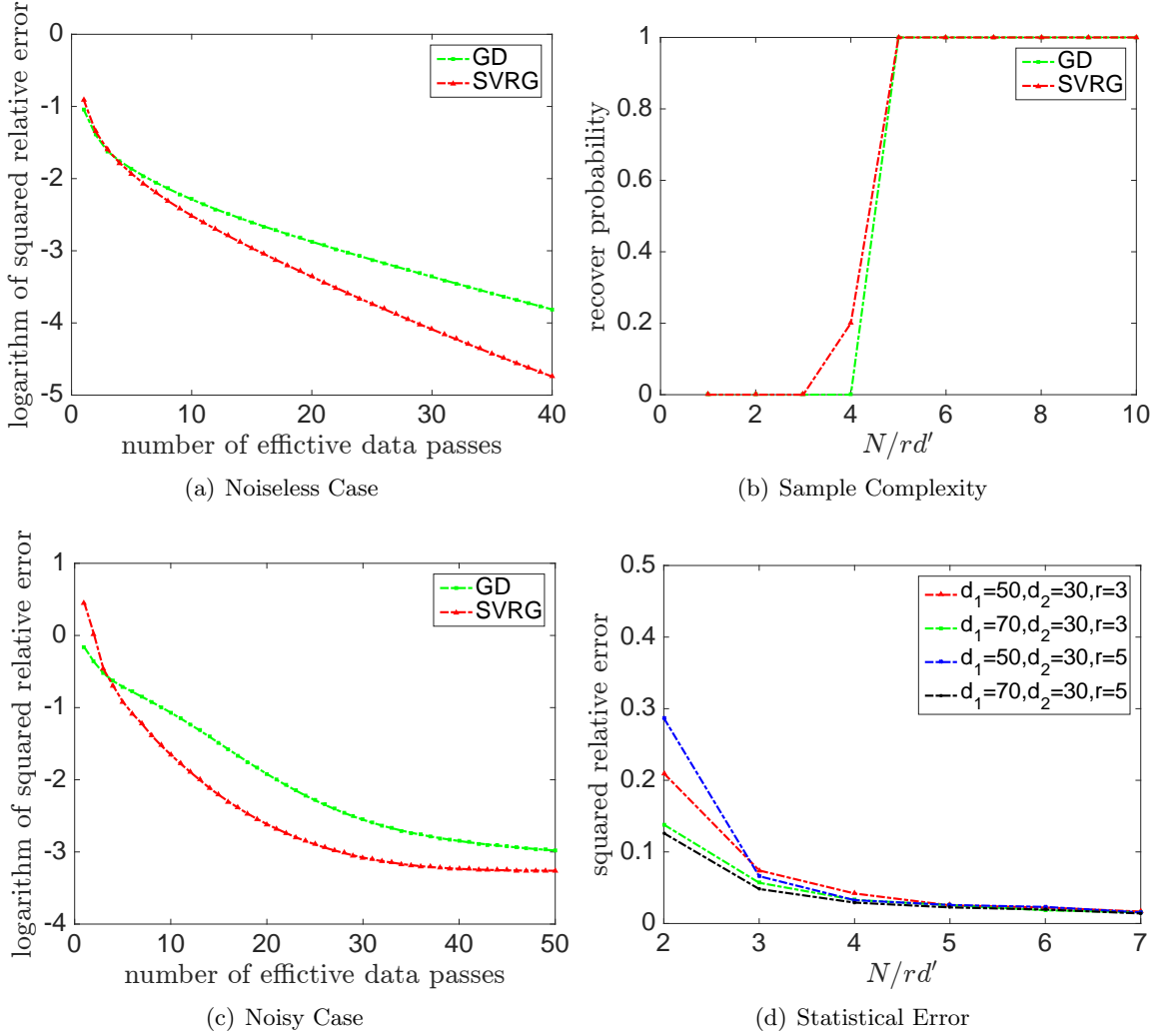

Figure 1: Experimental results for matrix sensing. (a) and (c) Rate of convergence for matrix sensing in the noiseless and noisy case respectively: squared relative error $\|\widehat{\mathbf{X}} - \mathbf{X}^*\|_F^2/\|\mathbf{X}^*\|_F^2$ in log scale versus number of effective data passes, which illustrate the linear rate of convergence of our algorithm and illustrates the improved performance of our method after the same number of effective data passes; (b) Probability of exact recovery versus rescaled sample size $N/(rd')$, which demonstrates the optimal sample complexity $N = O(rd')$; (d) Statistical error for matrix sensing: squared relative error $\|\widehat{\mathbf{X}} - \mathbf{X}^*\|_F^2/\|\mathbf{X}^*\|_F^2$ versus rescaled sample size $N/(rd')$, which is consistent with the statistical error rate.

*Proof of Lemma 5.1.* According to (5.1), we have

$$\langle \nabla \widetilde{f}(\mathbf{Z}), \mathbf{H} \rangle = \underbrace{\langle \nabla_{\mathbf{U}} \mathcal{L}(\mathbf{U}\mathbf{V}^\top), \mathbf{H}_U \rangle + \langle \nabla_{\mathbf{V}} \mathcal{L}(\mathbf{U}\mathbf{V}^\top), \mathbf{H}_V \rangle}_{I_1} + \frac{1}{2} \underbrace{\langle \widetilde{\mathbf{Z}} \widetilde{\mathbf{Z}}^\top \mathbf{Z}, \mathbf{H} \rangle}_{I_2}, \qquad (A.1)$$

where $\mathbf{H}_U, \mathbf{H}_V$ denote the top $d_1 \times r$ and bottom $d_2 \times r$ matrix of $\mathbf{H}$ respectively, and $\widetilde{\mathbf{Z}} = [\mathbf{U}; -\mathbf{V}]$.

In the following discussion, we are going to bound $I_1$ and $I_2$, respectively.

Recall $\mathbf{X}^* = \mathbf{U}^*\mathbf{V}^{*\top}$, and $\mathbf{X} = \mathbf{U}\mathbf{V}^\top$. Note $\nabla_\mathbf{U}\mathcal{L}(\mathbf{U}\mathbf{V}^\top) = \nabla\mathcal{L}(\mathbf{X})\mathbf{V}$ and $\nabla_\mathbf{V}\mathcal{L}(\mathbf{U}\mathbf{V}^\top) = \nabla\mathcal{L}(\mathbf{X})^\top \mathbf{U}$. Thus, for the term $I_1$, we have

$$\begin{aligned} I_1 &= \langle \nabla\mathcal{L}(\mathbf{X}), \mathbf{U}\mathbf{V}^\top - \mathbf{U}^*\mathbf{V}^{*\top} + \mathbf{H}_U \mathbf{H}_V^\top \rangle \\ &= \underbrace{\langle \nabla\bar{\mathcal{L}}(\mathbf{X}), \mathbf{X} - \mathbf{X}^* + \mathbf{H}_U \mathbf{H}_V^\top \rangle}_{I_{11}} + \underbrace{\langle \nabla\mathcal{L}(\mathbf{X}) - \nabla\bar{\mathcal{L}}(\mathbf{X}), \mathbf{X} - \mathbf{X}^* + \mathbf{H}_U \mathbf{H}_V^\top \rangle}_{I_{12}}, \end{aligned}$$

where $\nabla\mathcal{L}(\mathbf{X}) = \sum_{i=1}^N (\langle \mathbf{A}_i, \mathbf{X}\rangle - y_i)\mathbf{A}_i/N$, and $\nabla\bar{\mathcal{L}}(\mathbf{X}) = \sum_{i=1}^N \langle \mathbf{A}_i, \mathbf{X} - \mathbf{X}^*\rangle \mathbf{A}_i/N$. To begin with, we consider the term $I_{11}$. Recall that the linear measurement operator $\mathcal{A}$ satisfies the $4r$-RIP condition, and note that $\mathbf{X} - \mathbf{X}^*$ has rank at most $2r$, then we have

$$\langle \nabla\bar{\mathcal{L}}(\mathbf{X}), \mathbf{X} - \mathbf{X}^*\rangle = \frac{1}{N}\sum_{i=1}^N \langle \mathbf{A}_i, \mathbf{X} - \mathbf{X}^*\rangle^2 = \frac{1}{N}\|\mathcal{A}_N(\mathbf{X} - \mathbf{X}^*)\|_2^2 \geq (1 - \delta_{4r})\cdot \|\mathbf{X} - \mathbf{X}^*\|_F^2. \quad (A.2)$$

As for the remaining term in $I_{11}$, note that both $\mathbf{X} - \mathbf{X}^*$ and $\mathbf{H}_U \mathbf{H}_V^\top$ have rank at most $2r$, then we have

$$\begin{aligned} \left|\langle \nabla\bar{\mathcal{L}}(\mathbf{X}), \mathbf{H}_U \mathbf{H}_V^\top\rangle\right| &= \frac{1}{N}\left|\langle \mathcal{A}_N(\mathbf{X} - \mathbf{X}^*), \mathcal{A}_N(\mathbf{H}_U \mathbf{H}_V^\top)\rangle\right| \\ &\leq \delta_{4r}\cdot \|\mathbf{X} - \mathbf{X}^*\|_F \cdot \|\mathbf{H}_U \mathbf{H}_V^\top\|_F + \left|\langle \mathbf{X} - \mathbf{X}^*, \mathbf{H}_U \mathbf{H}_V^\top\rangle\right| \\ &\leq (1 + \delta_{4r})\cdot \|\mathbf{X} - \mathbf{X}^*\|_F \cdot \|\mathbf{H}_U \mathbf{H}_V^\top\|_F, \end{aligned} \quad (A.3)$$

where the first inequality follows from Lemma B.1 and the triangle inequality, and the second inequality holds because $|\langle \mathbf{A}, \mathbf{B}\rangle| \leq \|\mathbf{A}\|_F \cdot \|\mathbf{B}\|_F$. Therefore, combining (A.2) and (A.3), we obtain the lower bound of the term $I_{11}$

$$\begin{aligned} I_{11} &\geq (1 - \delta_{4r})\cdot \|\mathbf{X} - \mathbf{X}^*\|_F^2 - (1 + \delta_{4r})\cdot \|\mathbf{X} - \mathbf{X}^*\|_F \cdot \|\mathbf{H}_U \mathbf{H}_V^\top\|_F \\ &\geq \left(\frac{1}{2} - \frac{3}{2}\delta_{4r}\right)\cdot \|\mathbf{X} - \mathbf{X}^*\|_F^2 - \frac{1 + \delta_{4r}}{8}\cdot \|\mathbf{H}\|_F^4, \end{aligned} \quad (A.4)$$

where the second inequality holds because the inequality $2ab \leq a^2 + b^2$. Next, for the term $I_{12}$, we can get

$$\begin{aligned} \left|\langle \nabla\mathcal{L}(\mathbf{X}) - \nabla\bar{\mathcal{L}}(\mathbf{X}), \mathbf{X} - \mathbf{X}^*\rangle\right| &\leq \|\nabla\mathcal{L}(\mathbf{X}) - \nabla\bar{\mathcal{L}}(\mathbf{X})\|_2 \cdot \|\mathbf{X} - \mathbf{X}^*\|_* \\ &\leq \sqrt{2r}\cdot \left\|\frac{1}{N}\sum_{i=1}^N \epsilon_i \mathbf{A}_i\right\|_2 \cdot \|\mathbf{X} - \mathbf{X}^*\|_F, \end{aligned} \quad (A.5)$$

where the first inequality holds because the Von Neumann trace inequality, and the second one is due to the fact that $\text{rank}(\mathbf{X} - \mathbf{X}^*) \leq 2r$. Thus by similar techniques, for the remaining term in $I_{12}$,

we have

$$|\langle \nabla \mathcal{L}(\mathbf{X}) - \nabla \bar{\mathcal{L}}(\mathbf{X}), \mathbf{H}_U \mathbf{H}_V^\top \rangle| \leq \sqrt{2r} \cdot \left\| \frac{1}{N} \sum_{i=1}^{N} \epsilon_i \mathbf{A}_i \right\|_2 \cdot \|\mathbf{H}_U \mathbf{H}_V^\top\|_F. \quad (A.6)$$

Thus, combining (A.5) and (A.6), the term $I_{12}$ has the following lower bound

$$\begin{aligned} I_{12} &\geq -\sqrt{2r} \cdot \left\| \frac{1}{N} \sum_{i=1}^{N} \epsilon_i \mathbf{A}_i \right\|_2 \cdot \left( \|\mathbf{X} - \mathbf{X}^*\|_F + \frac{1}{2} \|\mathbf{H}\|_F^2 \right) \\ &\geq -\frac{1}{2} \delta_{4r} \|\mathbf{X} - \mathbf{X}^*\|_F^2 - \frac{\delta_{4r}}{8} \|\mathbf{H}\|_F^4 - \frac{2r}{\delta_{4r}} \cdot \left\| \frac{1}{N} \sum_{i=1}^{N} \epsilon_i \mathbf{A}_i \right\|_2^2, \end{aligned} \quad (A.7)$$

where the first inequality follows from $2\|\mathbf{AB}\|_F \leq \|\mathbf{A}\|_F^2 + \|\mathbf{B}\|_F^2$, and the last inequality comes from the inequality that $2ab \leq \beta a^2 + b^2/\beta$, for any $\beta > 0$. Therefore, combining (A.4) and (A.7), we can lower bound $I_1$ as follows

$$I_1 \geq \left( \frac{1}{2} - 2\delta_{4r} \right) \cdot \|\mathbf{X} - \mathbf{X}^*\|_F^2 - \frac{1 + 2\delta_{4r}}{8} \|\mathbf{H}\|_F^4 - \frac{2r}{\delta_{4r}} \cdot \left\| \frac{1}{N} \sum_{i=1}^{N} \epsilon_i \mathbf{A}_i \right\|_2^2. \quad (A.8)$$

On the other hand, according to lemma A.1, for the term $I_2$ we have

$$\begin{aligned} I_2 &\geq \frac{1}{2} \|\widetilde{\mathbf{Z}}^\top \mathbf{Z}\|_F^2 - \frac{1}{2} \|\widetilde{\mathbf{Z}}^\top \mathbf{Z}\|_F \cdot \|\mathbf{H}\|_F^2 \\ &\geq \frac{1}{4} \|\widetilde{\mathbf{Z}}^\top \mathbf{Z}\|_F^2 - \frac{1}{4} \|\mathbf{H}\|_F^4, \end{aligned} \quad (A.9)$$

where the last inequality follows from the inequality that $2ab \leq a^2 + b^2$. By plugging (A.8) and (A.9) into (A.1), we can obtain

$$\langle \nabla \widetilde{F}_n(\mathbf{Z}), \mathbf{H} \rangle \geq \left( \frac{1}{2} - 2\delta_{4r} \right) \cdot \|\mathbf{X} - \mathbf{X}^*\|_F^2 + \frac{1}{8} \|\widetilde{\mathbf{Z}}^\top \mathbf{Z}\|_F^2 - \frac{1 + \delta_{4r}}{4} \|\mathbf{H}\|_F^4 - \frac{2r}{\delta_{4r}} \cdot \left\| \frac{1}{N} \sum_{i=1}^{N} \epsilon_i \mathbf{A}_i \right\|_2^2, \quad (A.10)$$

Furthermore, let $\widetilde{\mathbf{Z}}^* = [\mathbf{U}^*; -\mathbf{V}^*]$, then we have

$$\begin{aligned} \|\widetilde{\mathbf{Z}}^\top \mathbf{Z}\|_F^2 &= \langle \mathbf{Z}\mathbf{Z}^\top - \mathbf{Z}^*\mathbf{Z}^{*\top}, \widetilde{\mathbf{Z}}\widetilde{\mathbf{Z}}^\top - \widetilde{\mathbf{Z}}^*\widetilde{\mathbf{Z}}^{*\top} \rangle + \langle \mathbf{Z}^*\mathbf{Z}^{*\top}, \widetilde{\mathbf{Z}}\widetilde{\mathbf{Z}}^\top \rangle + \langle \mathbf{Z}\mathbf{Z}^\top, \widetilde{\mathbf{Z}}^*\widetilde{\mathbf{Z}}^{*\top} \rangle \\ &\geq \langle \mathbf{Z}\mathbf{Z}^\top - \mathbf{Z}^*\mathbf{Z}^{*\top}, \widetilde{\mathbf{Z}}\widetilde{\mathbf{Z}}^\top - \widetilde{\mathbf{Z}}^*\widetilde{\mathbf{Z}}^{*\top} \rangle \\ &= \|\mathbf{U}\mathbf{U}^\top - \mathbf{U}^*\mathbf{U}^{*\top}\|_F^2 + \|\mathbf{V}\mathbf{V}^\top - \mathbf{V}^*\mathbf{V}^{*\top}\|_F^2 - 2\|\mathbf{U}\mathbf{V}^\top - \mathbf{U}^*\mathbf{V}^{*\top}\|_F^2, \end{aligned} \quad (A.11)$$

where the first equality holds since $\widetilde{\mathbf{Z}}^{*\top} \mathbf{Z}^* = 0$, and the last inequality follows from $\langle \mathbf{A}\mathbf{A}^\top, \mathbf{B}\mathbf{B}^\top \rangle = \|\mathbf{A}^\top \mathbf{B}\|_F^2 \geq 0$. Therefore, according to Lemma B.3, we have

$$4\|\mathbf{X} - \mathbf{X}^*\|_F^2 + \|\widetilde{\mathbf{Z}}^\top \mathbf{Z}\|_F^2 = \|\mathbf{Z}\mathbf{Z}^\top - \mathbf{Z}^*\mathbf{Z}^{*\top}\|_F^2 \geq 4(\sqrt{2} - 1)\sigma_r \|\mathbf{H}\|_F^2, \quad (A.12)$$

where the equality follows from (A.11), and the inequality holds because of Lemma B.3 and the fact that $\sigma_r^2(\mathbf{Z}^*) = 2\sigma_r$. Note that we assume $\delta_{4r} \leq 1/16$. Thus plugging (A.12) into (A.10), we have

$$\langle \nabla \widetilde{f}(\mathbf{Z}), \mathbf{H} \rangle \geq \frac{\sigma_r}{10}\|\mathbf{H}\|_F^2 + \frac{1}{8}\|\mathbf{X} - \mathbf{X}^*\|_F^2 + \frac{1}{16}\|\widetilde{\mathbf{Z}}^\top \mathbf{Z}\|_F^2 - \frac{1}{3}\|\mathbf{H}\|_F^4 - 32r\left\|\frac{1}{N}\sum_{i=1}^N \epsilon_i \mathbf{A}_i\right\|_2^2,$$

which completes the proof. $\square$

## A.2 Proof of Lemma 5.2

*Proof.* For any $i \in [n]$, consider the term $\mathbb{E}\|\nabla \widetilde{f}_i(\mathbf{Z})\|_F^2$. According to (5.2), we have

$$\begin{aligned}
\|\nabla \widetilde{f}_i(\mathbf{Z})\|_F^2 &\leq 2\|\nabla_\mathbf{U} \ell_i(\mathbf{UV}^\top)\|_F^2 + 2\|\nabla_\mathbf{V} \ell_i(\mathbf{UV}^\top)\|_F^2 + \frac{1}{2}\|\mathbf{U}^\top\mathbf{U} - \mathbf{V}^\top\mathbf{V}\|_F^2 \cdot (\|\mathbf{U}\|_2^2 + \|\mathbf{V}\|_2^2) \\
&\leq 2\underbrace{\|\nabla_\mathbf{U} \ell_i(\mathbf{UV}^\top)\|_F^2}_{I_1} + 2\underbrace{\|\nabla_\mathbf{V} \ell_i(\mathbf{UV}^\top)\|_F^2}_{I_2} + \|\mathbf{U}^\top\mathbf{U} - \mathbf{V}^\top\mathbf{V}\|_F^2 \cdot \|\mathbf{Z}\|_2^2, \quad (A.13)
\end{aligned}$$

where the first inequality is due to the inequalities that $\|\mathbf{A} + \mathbf{B}\|_F^2 \leq 2\|\mathbf{A}\|_F^2 + 2\|\mathbf{B}\|_F^2$ and $\|\mathbf{AB}\|_F \leq \|\mathbf{A}\|_2 \cdot \|\mathbf{B}\|_F$, and the second inequality holds because we have $\max\{\|\mathbf{U}\|_2, \|\mathbf{V}\|_2\} \leq \|\mathbf{Z}\|_2$. In the following discussion, we are going to upper bound terms $I_1$ and $I_2$, respectively.

Consider $I_1$ in (A.13) first. By plugging in the definition of $\ell_i(\mathbf{UV}^\top)$ in (3.3), we have

$$\begin{aligned}
I_1 &= \left\|\frac{1}{b}\sum_{j \in [b]} \left(\langle \mathbf{A}_{i_j}, \mathbf{X} - \mathbf{X}^*\rangle - \epsilon_{i_j}\right)\mathbf{A}_{i_j}\mathbf{V}\right\|_F^2 \\
&\leq 2\left\|\frac{1}{b}\sum_{j \in [b]} \langle \mathbf{A}_{i_j}, \mathbf{X} - \mathbf{X}^*\rangle \mathbf{A}_{i_j}\mathbf{V}\right\|_F^2 + 2r\left\|\frac{1}{b}\sum_{j \in [b]} \epsilon_{i_j}\mathbf{A}_{i_j}\right\|_2^2 \cdot \|\mathbf{V}\|_2^2, \quad (A.14)
\end{aligned}$$

where the inequality holds since $\|\mathbf{A} + \mathbf{B}\|_F^2 \leq 2\|\mathbf{A}\|_F^2 + 2\|\mathbf{B}\|_F^2$, $\|\mathbf{AB}\|_F \leq \|\mathbf{A}\|_2 \cdot \|\mathbf{B}\|_F$ and rank$(\mathbf{V}) \leq r$. As for the first term in the R.H.S. of (A.14), according to definition of Frobenius norm, we have

$$\begin{aligned}
\left\|\frac{1}{b}\sum_{j \in [b]} \langle \mathbf{A}_{i_j}, \mathbf{X} - \mathbf{X}^*\rangle \mathbf{A}_{i_j}\mathbf{V}\right\|_F &= \sup_{\mathbf{W} \in \mathbb{R}^{d_1 \times r},\ \|\mathbf{W}\|_F \leq 1} \left\langle \frac{1}{b}\sum_{j \in [b]} \langle \mathbf{A}_{i_j}, \mathbf{X} - \mathbf{X}^*\rangle \mathbf{A}_{i_j}, \mathbf{WV}^\top \right\rangle \\
&= \sup_{\mathbf{W} \in \mathbb{R}^{d_1 \times r},\ \|\mathbf{W}\|_F \leq 1} \frac{1}{b}\langle \mathcal{A}_b^i(\mathbf{X} - \mathbf{X}^*), \mathcal{A}_b^i(\mathbf{WV}^\top)\rangle \\
&\leq \sup_{\mathbf{W} \in \mathbb{R}^{d_1 \times r},\ \|\mathbf{W}\|_F \leq 1} (1 + \delta_{4r}') \cdot \|\mathbf{X} - \mathbf{X}^*\|_F \cdot \|\mathbf{WV}^\top\|_F \\
&\leq (1 + \delta_{4r}') \cdot \|\mathbf{X} - \mathbf{X}^*\|_F \cdot \|\mathbf{V}\|_2, \quad (A.15)
\end{aligned}$$

where the first inequality follows from Lemma B.1 and the fact that $\mathbf{X} - \mathbf{X}^*, \mathbf{WV}^\top$ have rank at most $2r$, and the second inequality holds because $\|\mathbf{AB}\|_F \leq \|\mathbf{A}\|_2 \cdot \|\mathbf{B}\|_F$. Thus combining (A.14) and (A.15), we obtain the upper bound of $I_1$

$$I_1 \leq 2(1 + \delta'_{4r})^2 \cdot \|\mathbf{X} - \mathbf{X}^*\|_F^2 \cdot \|\mathbf{V}\|_2^2 + 2r \left\| \frac{1}{b} \sum_{j \in [b]} \epsilon_{i_j} \mathbf{A}_{i_j} \right\|_2^2 \cdot \|\mathbf{V}\|_2^2. \tag{A.16}$$

Next, we consider $I_2$ in (A.13). Similarly, we have

$$I_2 \leq 2(1 + \delta'_{4r})^2 \cdot \|\mathbf{X} - \mathbf{X}^*\|_F^2 \cdot \|\mathbf{U}\|_2^2 + 2r \left\| \frac{1}{b} \sum_{j \in [b]} \epsilon_{i_j} \mathbf{A}_{i_j} \right\|_2^2 \cdot \|\mathbf{U}\|_2^2. \tag{A.17}$$

Therefore, plugging (A.16) and (A.17) into (A.13), we obtain

$$\left\| \nabla \widetilde{f}_i(\mathbf{Z}) \right\|_F^2 \leq \left( 8(1 + \delta'_{4r})^2 \cdot \|\mathbf{X} - \mathbf{X}^*\|_F^2 + \|\mathbf{U}^\top \mathbf{U} - \mathbf{V}^\top \mathbf{V}\|_F^2 \right) \cdot \|\mathbf{Z}\|_2^2 + 8r \left\| \frac{1}{b} \sum_{j \in [b]} \epsilon_{i_j} \mathbf{A}_{i_j} \right\|_2^2 \cdot \|\mathbf{Z}\|_2^2, \tag{A.18}$$

where the inequality holds because we have that $\max\{\|\mathbf{U}\|_2, \|\mathbf{V}\|_2\} \leq \|\mathbf{Z}\|_2$. Therefore, we complete the proof. $\square$

## B Auxiliary lemmas

In this section, we provide several auxiliary lemmas from existing work to make our proof self-contained.

**Lemma B.1.** (Candes, 2008) Suppose that $\mathcal{A}_M$ satisfies $2r$-RIP with constant $\delta_{2r}$. Then, for all matrices $\mathbf{X}, \mathbf{Y} \in \mathbb{R}^{d_1 \times d_2}$ of rank at most $r$, we have

$$\left| \frac{1}{M} \langle \mathcal{A}_M(\mathbf{X}), \mathcal{A}_M(\mathbf{Y}) \rangle - \langle \mathbf{X}, \mathbf{Y} \rangle \right| \leq \delta_{2r} \|\mathbf{X}\|_F \cdot \|\mathbf{Y}\|_F.$$

**Lemma B.2.** (Tu et al., 2015) Suppose $\mathbf{M}, \mathbf{M}' \in \mathbb{R}^{d_1 \times d_2}$ are two rank-$r$ matrices. Furthermore, suppose they have following singular value decomposition $\mathbf{M} = \mathbf{U} \boldsymbol{\Sigma} \mathbf{V}^\top$ and $\mathbf{M}' = \mathbf{U}' \boldsymbol{\Sigma}' \mathbf{V}'^\top$. If $\|\mathbf{M} - \mathbf{M}'\|_2 \leq \sigma_r(\mathbf{M})/2$, then we have

$$d^2\left( [\mathbf{U}'; \mathbf{V}'] \boldsymbol{\Sigma}'^{1/2}, [\mathbf{U}; \mathbf{V}] \boldsymbol{\Sigma}^{1/2} \right) \leq \frac{2}{\sqrt{2} - 1} \frac{\|\mathbf{M}' - \mathbf{M}\|_F^2}{\sigma_r(\mathbf{M})}.$$

**Lemma B.3.** (Tu et al., 2015) For any two matrices $\mathbf{Z}_1, \mathbf{Z}_2 \in \mathbb{R}^{(d_1+d_2) \times r}$, we have

$$d^2(\mathbf{Z}_1, \mathbf{Z}_2) \leq \frac{1}{2(\sqrt{2}-1)\sigma_r^2(\mathbf{Z}_2)} \cdot \|\mathbf{Z}_1 \mathbf{Z}_1^\top - \mathbf{Z}_2 \mathbf{Z}_2^\top\|_F^2.$$

**Lemma B.4.** (Tu et al., 2015) For any two matrices $\mathbf{Z}_1, \mathbf{Z}_2 \in \mathbb{R}^{(d_1+d_2) \times r}$ satisfying $d(\mathbf{Z}_1, \mathbf{Z}_2) \leq \|\mathbf{Z}_2\|_2/4$, we have

$$\|\mathbf{Z}_1\mathbf{Z}_1^\top - \mathbf{Z}_2\mathbf{Z}_2^\top\|_F \leq \frac{9}{4}\|\mathbf{Z}_2\|_2 \cdot d(\mathbf{Z}_1, \mathbf{Z}_2).$$